\documentclass{article}

\usepackage{microtype}
\usepackage{graphicx}
\usepackage{booktabs} 

\usepackage{hyperref}

\usepackage{xcolor}
\usepackage[accepted]{icml2021}

\usepackage{subcaption}
\usepackage{comment}
\usepackage{amsmath,amssymb,bm} 
\usepackage{enumitem}
\usepackage{makecell}
\usepackage{wrapfig}
\usepackage{multirow}

\graphicspath{{../figs/}}
\usepackage{amssymb}
\usepackage{amsmath,amsfonts}
\usepackage{amsopn,bm,mathtools}

\usepackage{nicefrac}       
\newcommand{\vct}[1]{\boldsymbol{#1}} 
\newcommand{\mat}[1]{\boldsymbol{#1}} 

\newcommand\abs[1]{\left\lvert#1\right\rvert}

\newcommand{\ProbOpr}[1]{\mathbb{#1}}

\newcommand{\expect}[2]{%
\ifthenelse{\equal{#2}{}}{\ProbOpr{E}_{#1}}
{\ifthenelse{\equal{#1}{}}{\ProbOpr{E}\left[#2\right]}{\ProbOpr{E}_{#1}\left[#2\right]}}} 
\newcommand{\var}[2]{%
\ifthenelse{\equal{#2}{}}{\ProbOpr{VAR}_{#1}}
{\ifthenelse{\equal{#1}{}}{\ProbOpr{VAR}\left[#2\right]}{\ProbOpr{VAR}_{#1}\left[#2\right]}}} 

\DeclareMathOperator{\argmax}{arg\,max}

\newcommand{\vb}{\vct{b}}

\newcommand{\vq}{\vct{q}}

\newcommand{\mM}{\mat{M}}

\newcommand{\mW}{\mat{W}}
\newcommand{\mX}{\mat{X}}
\newcommand{\mY}{\mat{Y}}

\usepackage{xspace}

\makeatletter
\DeclareRobustCommand\onedot{\futurelet\@let@token\@onedot}
\def\@onedot{\ifx\@let@token.\else.\null\fi\xspace}

\makeatother

\usepackage{mathtools}

\usepackage{listings}

\newcommand{\Qtot}{Q^{\text{tot}}}
\newcommand{\Qim}{Q^{\text{tot}}_{\text{aux}}}
\usepackage{xspace}  

\newcommand{\starcraft}{\textsc{StarCraft}\xspace}
\newcommand{\ourfullapproach}{\textbf{R}andomized \textbf{E}ntity-wise \textbf{F}actorization for \textbf{I}magined \textbf{L}earning\xspace}
\newcommand{\ourapproach}{\textbf{\textsc{REFIL}}\xspace}
\definecolor{darkgreen}{rgb}{0, .5, 0}

\newcommand{\eat}[1]{}

\usepackage[colorinlistoftodos,linecolor=gray,backgroundcolor=none,bordercolor=blue,textsize=tiny, textwidth=34mm
]{todonotes}
\definecolor{fscolor}{rgb}{1, 0, 0}

\definecolor{wbcolor}{rgb}{0.5, 1, 0.5}

\definecolor{model_fig_orange}{HTML}{F09800}
\definecolor{model_fig_blue}{HTML}{0074F0}

\usepackage{microtype}
\microtypesetup{final,tracking=true,kerning=true,spacing=true,factor=1100,stretch=10,shrink=10}
\newcommand{\ourtitle}{{Randomized Entity-wise Factorization for \\
Multi-Agent Reinforcement Learning}}

\newsavebox{\measurebox}

\icmltitlerunning{Randomized Entity-wise Factorization for MARL}

\begin{document}

\twocolumn[
\icmltitle{\ourtitle}



\icmlsetsymbol{equal}{*}

\begin{icmlauthorlist}
\icmlauthor{Shariq Iqbal}{usc}
\icmlauthor{Christian A. Schroeder de Witt}{ox}
\icmlauthor{Bei Peng}{ox}
\icmlauthor{Wendelin B\"{o}hmer}{del}
\icmlauthor{Shimon Whiteson}{ox}
\icmlauthor{Fei Sha}{usc,goog}
\end{icmlauthorlist}

\icmlaffiliation{usc}{Department of Computer Science, University of Southern California}
\icmlaffiliation{ox}{Department of Computer Science, University of Oxford}
\icmlaffiliation{del}{Department of Software Technology, Delft University of
Technology}
\icmlaffiliation{goog}{Google Research}

\icmlcorrespondingauthor{Shariq Iqbal}{shariqiq@usc.edu}

\icmlkeywords{Multi-Agent, Reinforcement Learning, MARL, Multi-Task}

\vskip 0.3in
]



\printAffiliationsAndNotice{}  

\begin{abstract}
Multi-agent settings in the real world often involve tasks with varying types and quantities of agents and non-agent entities; however, common patterns of behavior often emerge among these agents/entities.
Our method aims to leverage these commonalities by asking the question: ``What is the expected utility of each agent when only considering a randomly selected sub-group of its observed entities?''
By posing this counterfactual question, we can recognize state-action trajectories within sub-groups of entities that we may have encountered in another task and use what we learned in that task to inform our prediction in the current one.
We then reconstruct a prediction of the full returns as a combination of factors considering these disjoint groups of entities and train this ``randomly factorized" value function as an auxiliary objective for value-based multi-agent reinforcement learning.
By doing so, our model can recognize and leverage similarities across tasks to improve learning efficiency in a multi-task setting.
Our approach, \ourfullapproach~(\ourapproach), outperforms all strong baselines by a significant margin in challenging multi-task StarCraft micromanagement settings.
\end{abstract}

\section{Introduction}
\label{sIntro}
\begin{figure}[t]
    \vspace{-0.1 in}
    \centering
    \includegraphics[width=0.55\linewidth]{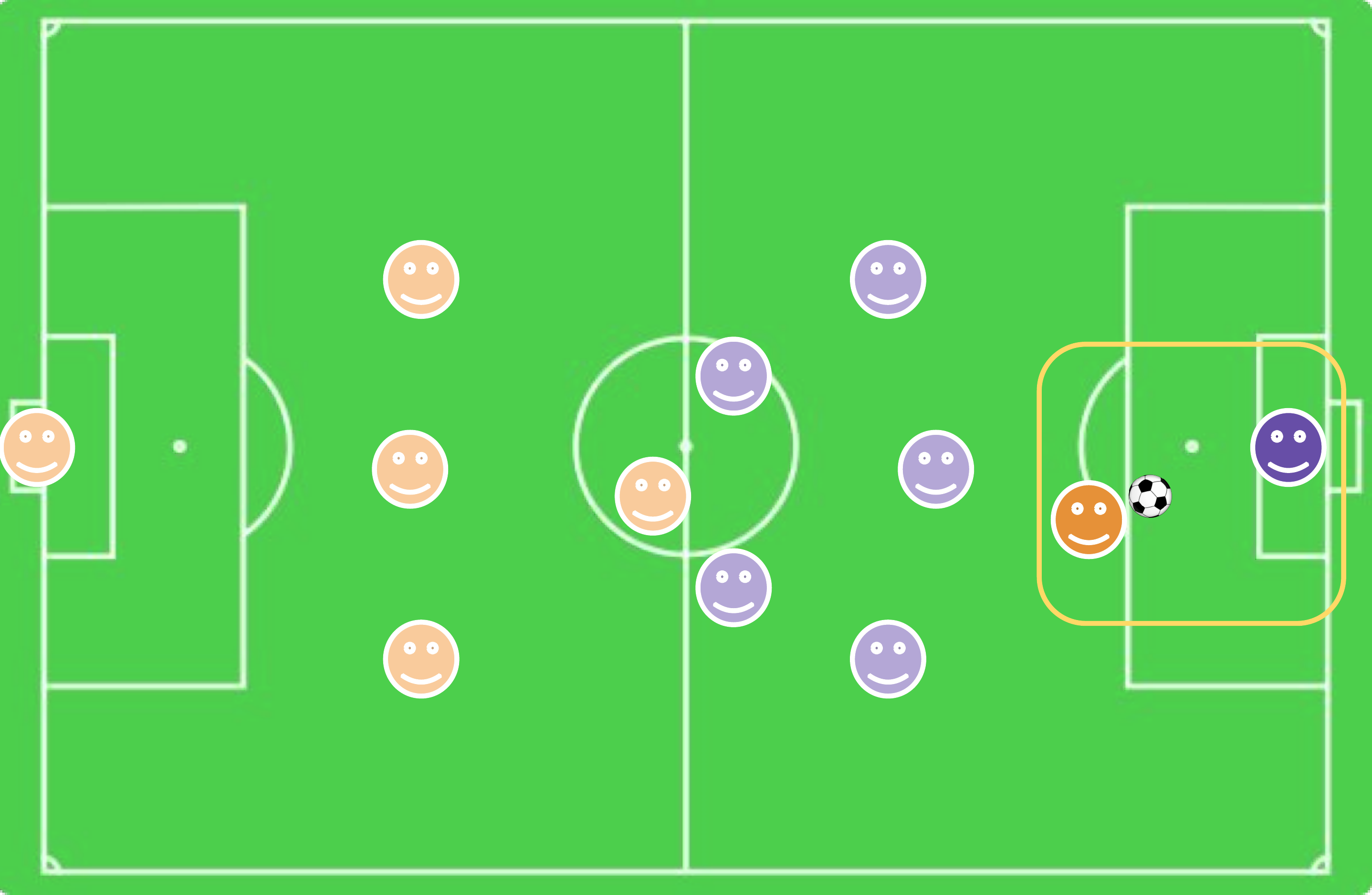}
    \vspace{-0.1 in}
    \caption{
    \footnotesize Breakaway sub-scenario in soccer. Agents in the yellow square can generalize this experience to similar subsequent experiences, regardless of the state of agents outside the square.}
    \label{fSoccerExample}
    \vspace{-0.25 in}
\end{figure}

Multi-agent reinforcement learning techniques often focus on learning in settings with fixed groups of agents and entities; however, many real-world multi-agent settings contain tasks across which an agent must deal with varying quantities and types of cooperative agents, antagonists, or other entities.
This variability in type and quantity of entities results in a combinatorial growth in the number of possible configurations, aggravating the challenge of learning control policies that generalize.
For example, the sport of soccer exists in many forms, from casual 5 vs.\ 5 to full scale 11 vs.\ 11 matches, with varying formations within each consisting of different quantities of player types (defenders, midfielders, forwards, etc.).
Within these varied tasks, however, exist common patterns.
For instance, a ``breakaway'' occurs in soccer when an attacker with the ball passes the defense and only needs to beat the goalkeeper in order to score (Figure~\ref{fSoccerExample}).
The goalkeeper and attacker can apply what they have learned in a breakaway to the next one, regardless of the task (e.g., 5 vs.\ 5).
If players can disentangle their understanding of common patterns from their surroundings, they should be able to learn more efficiently as well as share their experiences across \emph{all} forms of soccer.
These repeated patterns within sub-groups of entities can, in fact, be found in a wide variety of multi-agent tasks (e.g., heterogeneous swarm control~\citep{prorok2017impact} and StarCraft unit micromanagement~\citep{samvelyan2019starcraft}).
Our work\footnote{Code available at: \href{https://github.com/shariqiqbal2810/REFIL}{https://github.com/shariqiqbal2810/REFIL}} aims to develop a methodology for artificial agents to incorporate knowledge of these shared patterns to accelerate learning in a multi-task setting.

One way to leverage structural independence among agents, as in our soccer example, is to represent value functions as a combination of factors that depend on disjunct subsets of the state and action spaces~\citep{koller1999computing}.
These subsets are typically fixed in advance using domain knowledge and thus do not scale to complex domains where dependencies are unknown and may shift over time.
Recent approaches (e.g. VDN~\citep{sunehag2017value}, QMIX~\citep{rashid2018qmix}) in cooperative deep multi-agent reinforcement learning (MARL) factor value functions into separate components for each agent's action and observation space in order to enable decentralized execution.
These approaches learn a utility function for each agent that depends on the agent's own action and observations, resulting in a unique observation space for each task and exacerbating the challenge of learning in a multi-task setting.

\emph{How can we teach agents to be ``situationally aware'' of common patterns that are not pre-specified, such that they can share knowledge across tasks?}
Our main idea is as follows:
Given observed trajectories in a real task, we randomly partition entities into sub-groups to ``imagine'' that agents only observe a (random) subset of the entities they actually observe.
Then, in addition to estimating utility of their actions given the full observations, we use the same model to predict utilities in the imagined scenario, providing an opportunity to discover sub-group patterns that appear across tasks.
For example, we might sample a breakaway (or a superset of the breakaway entities) in both 5v5 soccer and 11v11, allowing our model to share value function factors across tasks.
We can then use these factors to construct a prediction of the full returns.

Of course, the possibility of sampling sub-groups that do not contain independent behavior exists.
Imagine a sub-group that looks like a breakaway, but in reality a defender is closing in on the attacker's left.
In such cases, we must include factors that account for the effect interactions outside of sampled sub-groups on each agent's utility.
Crucially however, the estimated utility derived from imagining a breakaway often provides at least \emph{some} information as to the agent's utility given the full observation (i.e., the agent knows there is value in dribbling toward the goal).
Imagined sub-group factors are combined with interaction factors to produce an estimate of the value function that we train as an auxiliary objective on top of a standard value function loss.
As such, our approach allows models to exploit shared inter-task patterns via factorization without losing any expressivity.
We emphasize that this approach does not rely on sampling an ``optimal'' sub-group.
In other words, there is no requirement to sample sub-groups that are independent from one another (c.f. Section \ref{ssGroupMatching}).
In fact, it is useful to learn a utility function for \emph{any} sub-group state that may appear in another task.

Our approach: \ourfullapproach~(\ourapproach) can be implemented easily in practice by using masks in attention-based models.
We evaluate our approach on complex StarCraft Multi-Agent Challenge (SMAC)~\citep{samvelyan2019starcraft} multi-task settings with varying agent teams, finding \ourapproach attains improved performance over state-of-the-art methods.


\section{Background and Preliminaries}
\label{sBackground}
In this work, we consider the \textit{decentralized partially observable Markov decision process} (Dec-POMDP)~\citep{oliehoek2016concise} with entities~\citep{NIPS2019_9184}, which describes fully cooperative multi-agent tasks.

\textbf{Dec-POMDPs with Entities}\ are described as tuples: $(\mathbf{S}, \mathbf{U}, \mathbf{O}, \mathit{P}, \mathit{r}, \mathcal{E}, \mathcal{A}, \Phi, \mu)$. $\mathcal{E}$ is the set of entities in the environment.
Each entity $e$ has a state representation $s^e$, and the global state is the set $\mathbf{s} = \{ s^e \vert e \in \mathcal{E} \} \in \mathbf{S}$.
Some entities can be agents $a \in \mathcal{A} \subseteq \mathcal{E}$.
Non-agent entities are parts of the environment that are not controlled by learning policies (e.g., landmarks, obstacles, agents with fixed behavior).
The state features of each entity consist of two parts: $s^e = [ f^e, \phi^e ]$ where $f^e$ represents the description of an entity's current state (e.g., position, orientation, velocity, etc.) while $\phi^e \in \Phi$ represents the entity's type (e.g., outfield player, goalkeeper, etc.), of which there is a discrete set.
An entity's type affects the state dynamics as well as the reward function and, importantly, it remains fixed for the duration of the entity's existence.
Not all entities may be visible to each agent, so we define a binary observability mask: $\mu(s^a, s^e) \in \{ 1, 0 \}$, where agents can always observe themselves $\mu(s^a, s^a) = 1, \forall a \in \mathcal{A}$.
Thus, an agent's observation is defined as $o^a = \{ s^e \vert \mu(s^a, s^e) = 1, e \in \mathcal{E} \} \in \mathbf{O} $.
Each agent $a$ can execute actions $u^a$, and the joint action of all agents is denoted $ \mathbf{u} = \{ u^a \vert a \in \mathcal{A} \} \in \mathbf{U}$.
$\mathit{P}$ is the state transition function which defines the probability $\mathit{P}(\mathbf{s}' | \mathbf{s}, \mathbf{u})$ and
$\mathit{r}(\mathbf{s}, \mathbf{u})$ is the reward function that maps the global state and joint actions to a single scalar reward.

Entities cannot be added during an episode, but they may become inactive (e.g.,\ a unit dying in StarCraft) and no longer affect transitions and rewards.
Since $\mathbf{s}$ and $\mathbf{u}$ are sets, their ordering does not matter, and our modeling construct should account for this (e.g., by modeling with permutation invariant/equivariant attention models~\citep{pmlr-v97-lee19d}).
In many domains, the set of entity types present $\{ \phi^e \vert e \in \mathcal{E} \}$ is fixed.
We are particularly interested in a multi-task setting where the quantity and types of entities are varied between tasks, as identifying patterns within sub-groups of entities is crucial to generalizing experience effectively in these cases.

\textbf{Learning}\ We aim to learn a set of policies that maximize expected discounted reward (returns).
$Q$-learning is specifically concerned with learning an accurate action-value function $\Qtot$ (defined below), and using this function to select the actions that maximize expected returns.
The optimal $Q$-function for our setting is defined as:
\vspace{-2mm}
\begin{align}
	\nonumber
    \Qtot(\mathbf s, \mathbf{u}) :=& \\
    \mathbb{E}\Big[
        \,{\textstyle\sum\nolimits_{t=0}^\infty} \gamma^t & \, r(\mathbf s_t, \mathbf u_t)
        \,\Big|\! \begin{array}{c}
            \scriptstyle
                \mathbf s_0 = \mathbf s,~ \mathbf u_0 = \mathbf u,~ 
                \mathbf s_{t+1} \sim P(\cdot|\mathbf s_t, \mathbf u_t) 
            \\[-0.5mm] \scriptstyle
                \mathbf u_{t+1} = \argmax 
                    \Qtot(\mathbf s_{t+1}, \cdot)
    \end{array}\!\Big] \\
    \nonumber
    =& \, r(\mathbf s, \mathbf u) +
    \gamma \, \mathbb E\Big[ 
        \max \Qtot(\mathbf s', \cdot) \,|\, 
        {\scriptstyle s' \sim P(\cdot|\mathbf s, \mathbf u)}
    \Big] \,
\end{align}
Partial observability is typically handled by using the history of actions and observations as a proxy for state, often processed by a recurrent neural network~\citep{Hausknecht15}: 
$\Qtot_\theta(\bm{\tau}_t, \mathbf u_t) \approx \Qtot(\mathbf s_t, \mathbf u_t)$
where the trajectory is $\tau_t^a:=(o^a_0, u^a_0, \ldots, o^a_t)$ and $\bm{\tau}_t := \{\tau^a_t\}_{a \in\mathcal A}$.
To learn the $Q$-function, deep reinforcement learning uses neural networks as function approximators trained to minimize the loss function:\!\!
\begin{align} \label{eq:dqn_loss}
    \nonumber
    \mathcal{L}_Q(\theta) :=& \, \mathbb{E}\Big[\! 
        \Big(\! y^\text{tot}_t
            \!- \Qtot_\theta(\bm\tau_t, \mathbf u_t) \!\Big)^{\!2}
        \Big| (\bm\tau_t, \mathbf u_t, r_t, \bm\tau_{t+1}) \sim \mathcal D    
    \Big] \\
    y^\text{tot}_t :=& \, r_t + \gamma \Qtot_{\bar\theta}\big(\bm\tau_{t+1}, 
                \argmax \Qtot_\theta(\bm\tau_{t+1}, \, \cdot \, )\big) 
\vspace{-2mm}
\end{align}
where $\bar\theta$ are the parameters of a target network that is copied from $\theta$ periodically to improve stability~\citep{mnih2015human} and $\mathcal D$ is a replay buffer~\citep{Lin92} that stores transitions collected by an exploratory policy (typically $\epsilon$-greedy).
Double deep $Q$-learning \citep{Hasselt16} mitigates overestimation 
of the learned values by using actions that 
maximize $\Qtot_\theta$ as inputs for the target network $\Qtot_{\bar\theta}$.

\textbf{Value Function Factorization} Centralized training for decentralized execution (CTDE) has been a major focus in recent efforts in deep multi-agent RL~\citep{lowe2017multi,Foerster2017-do,sunehag2017value,rashid2018qmix,pmlr-v97-iqbal19a}.
Some methods achieve CTDE through factoring $Q$-functions into monotonic combinations of per-agent utilities, with each depending only on a single agent's history of actions and observations $Q^a(\tau^a, u^a)$.
This factorization allows agents to independently maximize their local utility functions in a decentralized manner with their selected actions combining to form the optimal joint action.
While factored value functions can only represent a limited subset of all possible value functions \citep{boehmer2019dcg}, they tend to perform better empirically than those that learn unfactored joint action value functions~\citep{oliehoek2008exploiting}.

QMIX~\citep{rashid2018qmix} improves over value decomposition networks (VDN)~\citep{sunehag2017value} by using a more expressive factorization than a summation of factors:
\[\Qtot = g\big(Q^1(\tau^1, u^1; \theta_Q), \dots, Q^{\vert \mathcal{A} \vert}(\tau^{\vert \mathcal{A} \vert}, u^{\vert \mathcal{A} \vert}; \theta_Q); \theta_g\big)\]
The parameters of the monotonic mixing function $\theta_g$ are generated by a hyper-network \citep{ha2016hypernetworks} conditioning on the global state $\mathbf{s}$: $\theta_g = h(\mathbf{s}; \theta_h)$.
Every state can therefore have a different mixing function; however, the mixing functions's monotonicity maintains decentralizability,
as agents can greedily maximize $\Qtot$ without communication.
All parameters $\theta = \{ \theta_Q, \theta_h \}$ are trained with the DQN loss of Equation~\ref{eq:dqn_loss}.

\textbf{Attention Mechanisms for MARL}
Attention models have recently generated intense interest due to their ability to incorporate information across large contexts, including in  MARL~\citep{jiang2018learning,pmlr-v97-iqbal19a,Long2020Evolutionary}.
Importantly for our purposes, they can process variable sized sets of fixed length vectors (in our case entities).
At the core of these models is a parameterized transformation known as multi-head attention~\citep{vaswani2017attention} that allows entities to selectively extract information from other entities based on their local context.

We define $\mX$ as a matrix where each row corresponds to the state representation (or its transformation) of an entity.
The global state $\mathbf{s}$ is represented in matrix form as $\mX^{\mathcal{E}}$ where $\mX_{e,*} = s^e$.
Our models consist of entity-wise feedforward layers $\text{eFF}(\mX)$, which  apply an identical linear transformation to all input entities and multi-head attention layers $\text{MHA}\left( \mathcal{A}, \mX, \mat{M} \right)$, which integrate information across entities.
The latter take three arguments: the set of agents $\mathcal{A}$ for which to compute an output vector, the matrix $\mX \in \mathbb{R}^{\vert \mathcal{E} \vert \times d}$ where $d$ is the dimensionality of the input representations, and a mask $\mat{M} \in \mathbb{R}^{\vert \mathcal{A} \vert \times \vert \mathcal{E} \vert}$.
The layer outputs a matrix $\mat{H} \in \mathbb{R}^{\vert \mathcal{A} \vert \times h}$ where $h$ is the hidden dimension of the layer.
The row $\mat{H}_{a,*}$ corresponds to a weighted sum of linearly transformed representations from all entities selected by agent $a$.
Importantly, if the entry of the mask $\mat{M}_{a, e} = 0$, then entity $e$'s representation is not included in $\mat{H}_{a, *}$.
Masking enables 1) decentralized execution by providing the mask $\mat{M}^{\mu}_{a, e} = \mu(s^a, s^e)$, such that agents can only see entities observable by them in the environment, and 2) ``imagination'' of the returns among sub-groups of entities.
We integrate entity-wise feedforward layers and multi-head attention into QMIX in order to share a model across tasks where the number of agents and entities is variable and build our approach from there.
The exact process of computing attention layers, as well as the specifics of our attention-augmented version of QMIX are described in detail in the Supplement.

\vspace{-0.1 in}
\section{\ourapproach}
\label{sImagine}

\begin{figure*}[t]
    \centering
    \centerline{\includegraphics[width=0.8\linewidth]{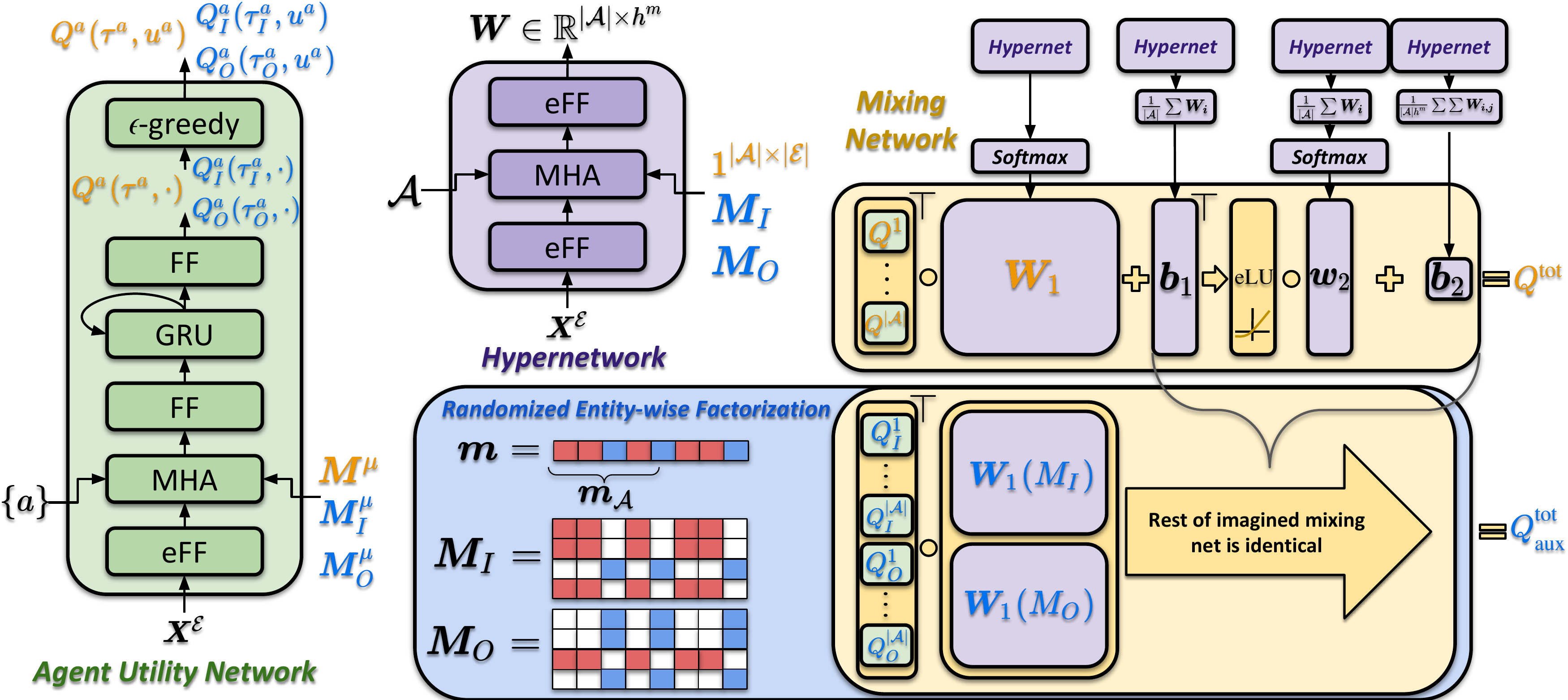}}
    \caption{
        \footnotesize
        Schematic for \ourapproach. Values colored \textcolor{model_fig_orange}{orange} or \textcolor{model_fig_blue}{blue} are used for computing \textcolor{model_fig_orange}{$\Qtot$} and \textcolor{model_fig_blue}{$\Qim$} respectively.
        \textbf{(left)} Agent-specific utility networks.
        These are decentralizable due to the use of an observability mask ($\mM^\mu$).
        We include Gated Recurrent Units~\citep{chung2014empirical} to retain information across timesteps in order to handle partial observability.
        \textbf{(top center)} Hypernetworks used to generate weights for the mixing network.
        We use a softmax function on the weights across the hidden dimension to enforce non-negativity, which we find empirically to be more stable than the standard absolute value function.
        Hypernetworks are not restricted by partial observability since they are only required during training and not execution.
        \textbf{(top right)} The mixing network used to calculate $\Qtot$.
        \textbf{(bottom right)} Procedure for performing randomized entity-wise factorization.
        For masks $M_I$ and $M_O$, colored spaces indicate a value of 1 (i.e., the agent designated by the row will be able to see the entity designated by the column), while white spaces indicate a value of 0.
        The color indicates which group the entity belongs to, so agents in the red group see red entities in $M_I$ and blue entities in $M_O$.
        Agents are split into sub-groups and their utilities are calculated for both interactions within their group, as well as to account for the interactions outside of their group, then monotonically mixed to predict \textcolor{model_fig_blue}{$\Qim$}.
    }
    \label{fModelDiagram}
    \vspace{-0.2 in}
\end{figure*}

We now propose \ourfullapproach~(\ourapproach).
We observe that common patterns often emerge in sub-groups of entities within complex multi-agent tasks (cf.\ soccer breakaway example in \S\ref{sIntro}) and hypothesize that learning to predict agents' utilities within sub-groups of entities is a strong inductive bias that allows models to share information more freely across tasks.
We instantiate our approach by constructing an estimate of the value function from factors based on randomized sub-groups, sharing parameters with the full value function, and training this factorized version of the value function as an auxiliary objective.

\subsection{Main Idea}

\textbf{Random Partitioning}\ Given an episode sampled from a replay buffer, we first randomly partition all entities in $\mathcal E$ into two disjunct groups, held fixed for the episode.
We denote the partition by a random binary\footnote{We first draw $p \in (0,1)$ uniformly, followed by $\vert \mathcal{E} \vert$ independent draws from a Bernoulli($p$) distribution. Partitioning into two groups induces a uniform distribution over all possible sub-groups.
} vector $\vct m \in \{0,1\}^{\vert \mathcal{E} \vert}$.
$m_e$ indicates whether entity $e$ is in the first group. The negation $\neg m_e$ represents whether $e$ is in the second group. The subset of all agents is denoted $\vct m_{\!\mathcal A} := [m_a]_{a \in \mathcal A}$.
With these vectors, we construct binary attention masks $\mM_I$ and $\mM_O$:
\begin{equation}
\label{eMasks}
    \mat{M}_I := \vct{m}_{\!\mathcal A} \, \vct{m}^\top 
                  \vee \neg\vct{m}_{\!\mathcal A} \neg\vct{m}^\top,
    \mat{M}_O := \neg \mat{M}_I.
\end{equation}
where $\mM^{\mu}_I[a, e]$ indicates whether agent $a$ and entity $e$ are in the same group, and $\mM^{\mu}_O[a, e]$ indicates the opposite.
They are further combined with a partial observability mask $\mM^{\mu}$, which is provided by the environment, to generate the final attention masks
\begin{equation}
    \label{eMasks2}
    \mM^{\mu}_I := \mM^{\mu} \wedge \mat{M}_I \,,
    \mM^{\mu}_O := \mM^{\mu} \wedge \mat{M}_O 
\end{equation}
These matrices are of size $\vert \mathcal{A} \vert \times \vert \mathcal{E} \vert$ and will be used by the multi-head attention layers to constrain which entities can be observed by agents.

\textbf{Counterfactual Reasoning}\ Given an \emph{imagined} partition $\vct{m}$, an agent $a$ can examine its history of observations and actions and reason \emph{counterfactually} what its utility would be had it \emph{solely} observed the entities in its group.
We call this quantity \emph{in-group utility} and denote it by $Q^a_I(\tau^a_I, u^a; \theta_Q)$.
In order to account for the potential interactions with entities outside of the agents group, we calculate an \emph{out-group utility}: $Q^a_O(\tau^a_O, u^a; \theta_Q)$.
Note that the real and imagined utilities share the same parameters $\theta_Q$, allowing us to leverage imagined experience to improve utility prediction in real scenarios and vice versa.
Breaking the fully observed utilities $Q^a$ into these randomized sub-group factors is akin to breaking an image into cut-outs of the comprising entities.
While the ``images'' (i.e. states) from each task are a unique set, it’s likely that the pieces comprising them share similarities.

Since we do not know the returns within the imagined sub-groups, we must ground our predictions in the observed returns.
Just as QMIX learns a value function with $n$ factors ($Q^a$ for each agent), we learn an imagined value function with $2n$ factors ($Q_I^a$ and $Q_O^a$ for each agent) that estimates the same value:
\begin{align}
\Qtot &= g\big(Q^1 , \dots, Q^{\vert \mathcal{A} \vert}; h(\mathbf{s}; \theta_h, \mM )\big)\nonumber\\
& \approx \Qim = g\big(Q^1_I, \dots, Q^{\vert \mathcal{A} \vert}_I, Q^1_O, \dots, Q^{\vert \mathcal{A} \vert}_O; \nonumber\\
& \hspace{4em} h(\mathbf{s}; \theta_h, \mM_I), h(\mathbf{s}; \theta_h, \mM_O )\big)
\label{eQMatch}
\end{align}
Where $g(\cdot)$ are mixing networks whose parameters are generated by hypernetworks $h(\mathbf{s}; \theta_h, \mM)$.
This network's first layer typically takes $n$ inputs, one for each agent.
Since we have $2n$ factors, we simply concatenate two generated versions of the input layer (using $\mM_I$ and $\mM_O$).
We then apply the network to the concatenated utilities
$Q^a_I(\tau^a_I, u^a)$ and $Q^a_O(\tau^a_O, u^a)$ of all agents $a$, 
to compute the predicted value $\Qim$.
This procedure is visualized in Figure~\ref{fModelDiagram} and described in more detail in the Supplement.

\begin{figure*}[t]
    \centering
    \begin{subfigure}[t]{0.48\textwidth}
        \centerline{\includegraphics[width=\linewidth]{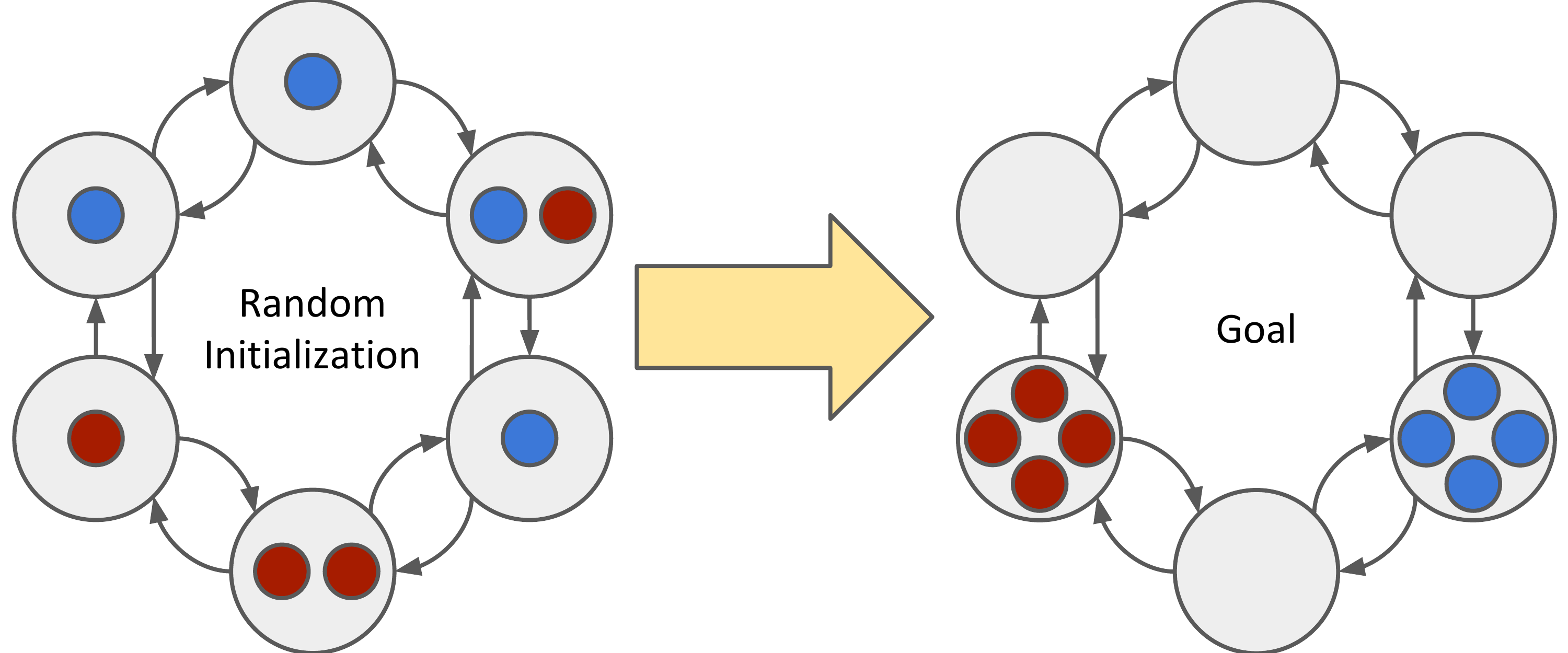}}
        \caption{Game Visualization}
        \label{fGameVis}
    \end{subfigure}
    \hspace{0.5 in}
    \begin{subfigure}[t]{0.29\textwidth}
        \centerline{\includegraphics[width=\linewidth]{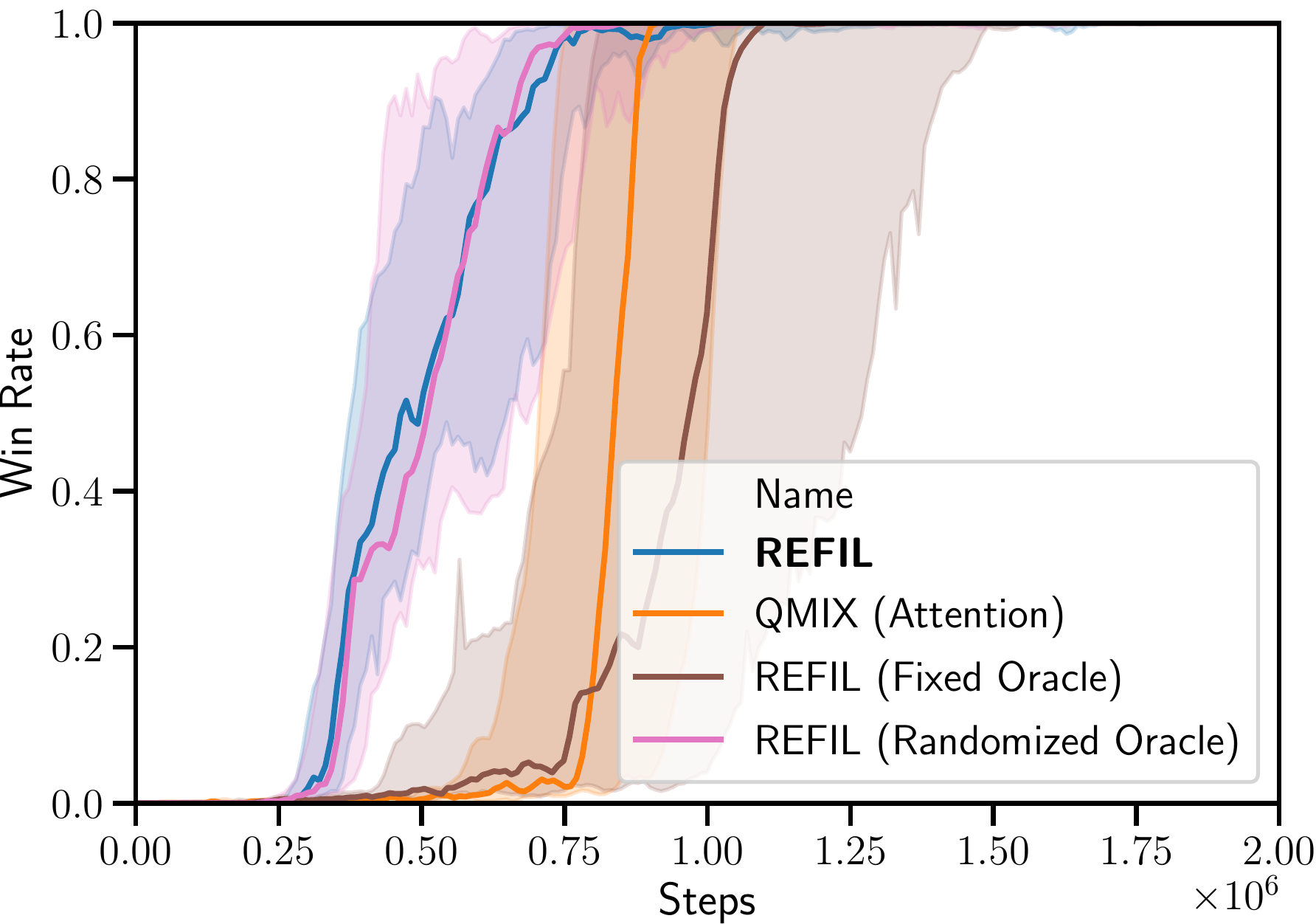}}
        \caption{Win Rate over Time}
        \label{fGame_win_rate}
    \end{subfigure}
    \caption{\small
             Group Matching Game. We use the values $n_a = 8$, $n_c = 6$, and $n_g = 2$ in our experiments.
             Shaded region is a 95\% confidence interval across 24 runs.
        }
    \label{fGame_results}
    \vspace{-0.1 in}
\end{figure*}

Importantly, since the mixing network is generated by the \emph{full state context}, our model can weight factors contextually.
For example, if the agent $a$'s sampled sub-group contains all relevant information to compute its utility such that $Q_I^a \approx Q^a$, then the mixing networks can weight $Q_I^a$ more heavily than $Q_O^a$.
Otherwise, the networks learn to balance $Q_I^a$ and $Q_O^a$ for each agent, in order to estimate $\Qtot$.
In this way, we can share knowledge in similar sub-group states across tasks while accounting for the differences in utility that result from the out-of-group context.

\textbf{Learning}\ We now describe the overall learning objective of \ourapproach.
To enforce Equation~\ref{eQMatch}, we replace $\Qtot$ in Equation~\ref{eq:dqn_loss} with $\Qim$, resulting a new loss $\mathcal{L}_\text{aux}$.
We combine the standard QMIX loss (Eq.~\ref{eq:dqn_loss}) $\mathcal{L}_Q$ with this auxiliary loss to form:
\begin{equation}
\mathcal{L} = (1-\lambda) \mathcal{L}_Q + \lambda \mathbb{E}_{\vct{m}}\mathcal{L}_\text{aux}
\end{equation}
where $\lambda$ controls the tradeoff between the two losses.
Note that we randomly partition in each episode, hence the expectation with respect to the partition vector $\vct{m}$.
We emphasize that the sub-groups are \emph{imagined}.
While we compute $\Qim$ and its related quantities, we do not use them to select actions in Equation~\ref{eq:dqn_loss}.
Action selection is performed by each agent maximizing $Q^a$ given their local observations.
This greedy local action selection is guaranteed to maximize $\Qtot$ due to the monotonic structure of the mixing network~\citep{rashid2018qmix}.
Moreover, our auxiliary objective is only used in training, and execution in the environment does not use random factorization.
Treating random factorization as an auxiliary task, rather than as a representational constraint, allows us to retain the expressivity of QMIX value functions (without sub-group factorization) while exploiting the existence of shared sub-group states across tasks.

\subsection{Implementation Details}

The model architecture is shown in Figure~\ref{fModelDiagram}, with more details described in the supplement.
``Imagination'' can be implemented efficiently using attention masks. Specifically two additional passes through the network are needed, with $\mM^{\mu}_O$ and $\mM^{\mu}_I$ as masks instead of $\mM^{\mu}$, per training step.
These additional passes can be parallelized by computing all necessary quantities in one batch on GPU.
It is feasible to split entities into an arbitrary number $i$ of random sub-groups without using more computation by sampling several disjunct vectors $\vct{m}^i$ and combining them them in the same way as we combine $\vct{m}$ and $\neg \vct{m}$ in Equation~\ref{eMasks} to form $\mM_I$ and $\mM_O$. Doing so could potentially bias agents towards considering patterns within smaller subsets of entities.

\section{Experimental Results}
\label{sExperiments}
\vspace{-0.1 in}
In our experiments, we aim to answer the following questions: 1) Are randomized counterfactuals an efficient means for leveraging common patterns? 2) Does our approach improve generalization in a multi-task setting? 3) Is training as an auxiliary objective justified?
We begin with experiments in a simple domain we construct such that agents' decisions rely only on a \emph{known subset} of all entities, so we can compare our approach to those that use this domain knowledge.
Then, we move on to testing on complex StarCraft micromanagement tasks to demonstrate our method's ability to scale to complex domains.

\subsection{Group Matching Game}
\label{ssGroupMatching}
In order to answer our first question, we construct a group matching game, pictured in Figure~\ref{fGameVis}, where each agent only needs to consider a subset of other agents to act effectively and we know that subset as ground truth (unlike in more complex domains such as StarCraft).
Agents (of which there are $n_a$) are randomly placed in one of $n_c$ cells and assigned to one of $n_g$ groups (represented by the different colors) at the start of each episode.
Each unique group assignment corresponds to a task.
Agents can choose from three actions: move clockwise, stay, and move counter-clockwise.
Their ultimate goal is to be located in the same cell as the rest of their group members, at which point an episode ends.
There is no restriction on which cell agents form a group in (e.g., both groups can form in the same cell).
All agents share a reward of 2.5 when any group is completed (and an equivalent penalty for a formed group breaking) as well as a penalty of -0.1 for each time step in order to encourage agents to solve the task as quickly as possible.
Agents' entity-state descriptions $s^e$ include the cell that the agent is currently occupying as well as the group it belongs to (both one-hot encoded), and the task is fully-observable.
Notably, agents can act optimally while only considering a subset of observed entities.

Ground-truth knowledge of relevant entities enables us to disentangle two aspects of our approach: the use of entity-wise factorization in general and specifically using randomly selected factors.
We would like to answer the question: does our method rely on sampling the ``right'' groups of entities (i.e., those with no interactions between them), or is the randomness of our method a feature that promotes generalization?
We construct two approaches that use this knowledge to build factoring masks $M_I$ and $M_O$ that are used in place of randomly sampled groups (otherwise the methods are identical to \ourapproach).
\textit{\ourapproach (Fixed Oracle)} directly uses the ground truth group assignments (different for each task) to build masks.
\textit{\ourapproach (Randomized Oracle)} randomly samples sub-groups from the ground truth groups only, rather than from all possible entities.
We additionally train \ourapproach and \textit{QMIX (Attention)} (i.e., \ourapproach with no auxiliary loss).

Figure \ref{fGame_win_rate} shows that using domain knowledge does not significantly improve performance in this domain (\textit{QMIX (Attention)} vs.\ \textit{\ourapproach (Fixed Oracle)}).
In fact our \emph{randomized} factorization approach outperforms the use of domain knowledge.
The randomization in \ourapproach therefore appears to be crucial.
Our hypothesis is that randomization of sub-group factors enables better knowledge sharing across tasks.
For example, the situation where two agents from the same group are located in adjacent cells occurs within \emph{all} possible group assignments.
When sampling randomly, our approach occasionally samples these two agents alone in their own group.
Even if the rest of the context in a given episode has never been seen  before, as long as this sub-scenario has been seen, the model has some indication of the value associated with each action.
Even when restricting the set of entities to form sub-groups with those that we know can be relevant to each agent (\textit{\ourapproach (Randomized Oracle)}) we find that performance does not significantly improve.
These results suggest that randomized sub-group formation for \ourapproach is a viable strategy, and the main benefit of our approach is to promote generalization across tasks by breaking value function predictions into reusable components, even when the sampled sub-groups are not completely independent.

\vspace{-0.1 in}
\subsection{\starcraft}
\vspace{-0.1 in}
We next test on the StarCraft multi-agent challenge (SMAC) ~\citep{samvelyan2019starcraft}.
The tasks in SMAC involve micromanagement of units in order to defeat a set of enemy units in battle.
Specifically, we consider a multi-task setting where we train our models simultaneously on tasks with variable types and quantities of agents.
We hypothesize that our approach is especially beneficial in this setting, as it should encourage models to learn utilities for common patterns and generalize to more diverse settings as a result.
The dynamic setting involves minor modifications to SMAC but we change the environment as little as possible to maintain the challenging nature of the tasks.
In the standard version of SMAC, both state and action spaces depend on a fixed number of agents and enemies, so our modifications, discussed in detail in the supplement, alleviate these problems.

\begin{figure*}[t]
    \centering
    \begin{subfigure}[t]{0.333\textwidth}
        \centerline{\includegraphics[width=\linewidth]{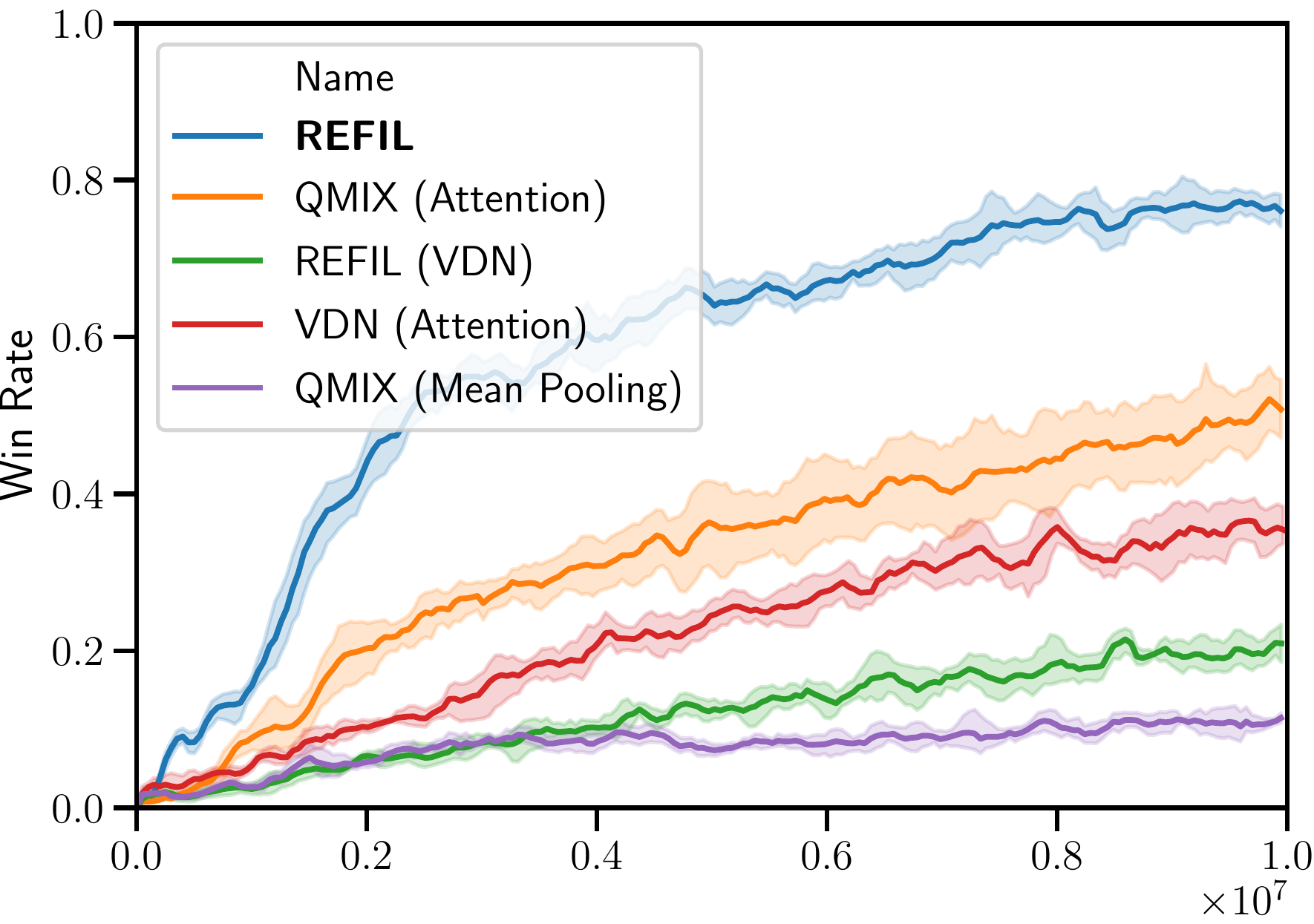}}
        \label{fSC_3-8sz_win_rate}
    \end{subfigure}
    \begin{subfigure}[t]{0.32\textwidth}
        \centerline{\includegraphics[width=\linewidth]{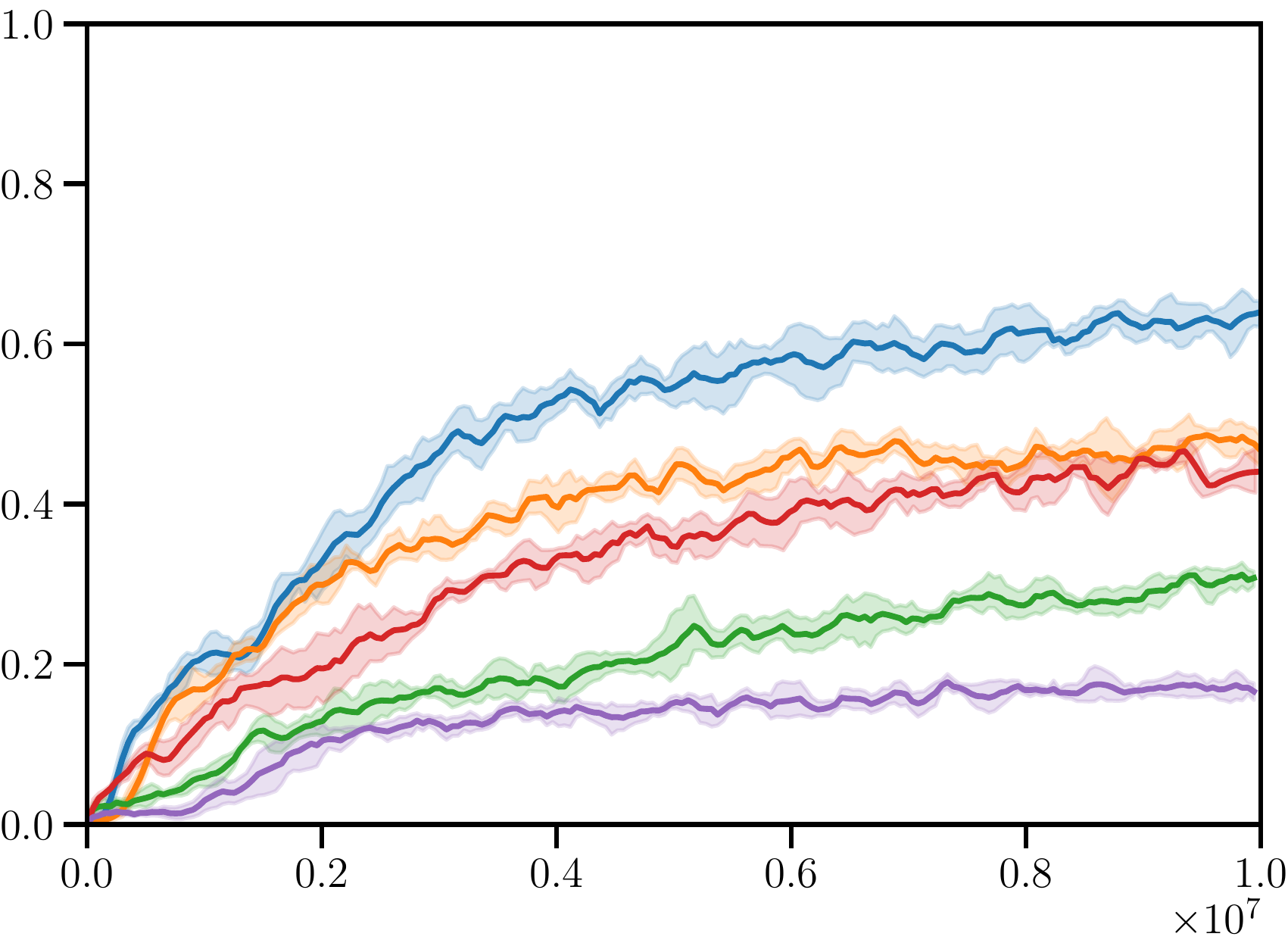}}
        \label{fSC_3-8csz_win_rate}
    \end{subfigure}
    \begin{subfigure}[t]{0.32\textwidth}
        \centerline{\includegraphics[width=\linewidth]{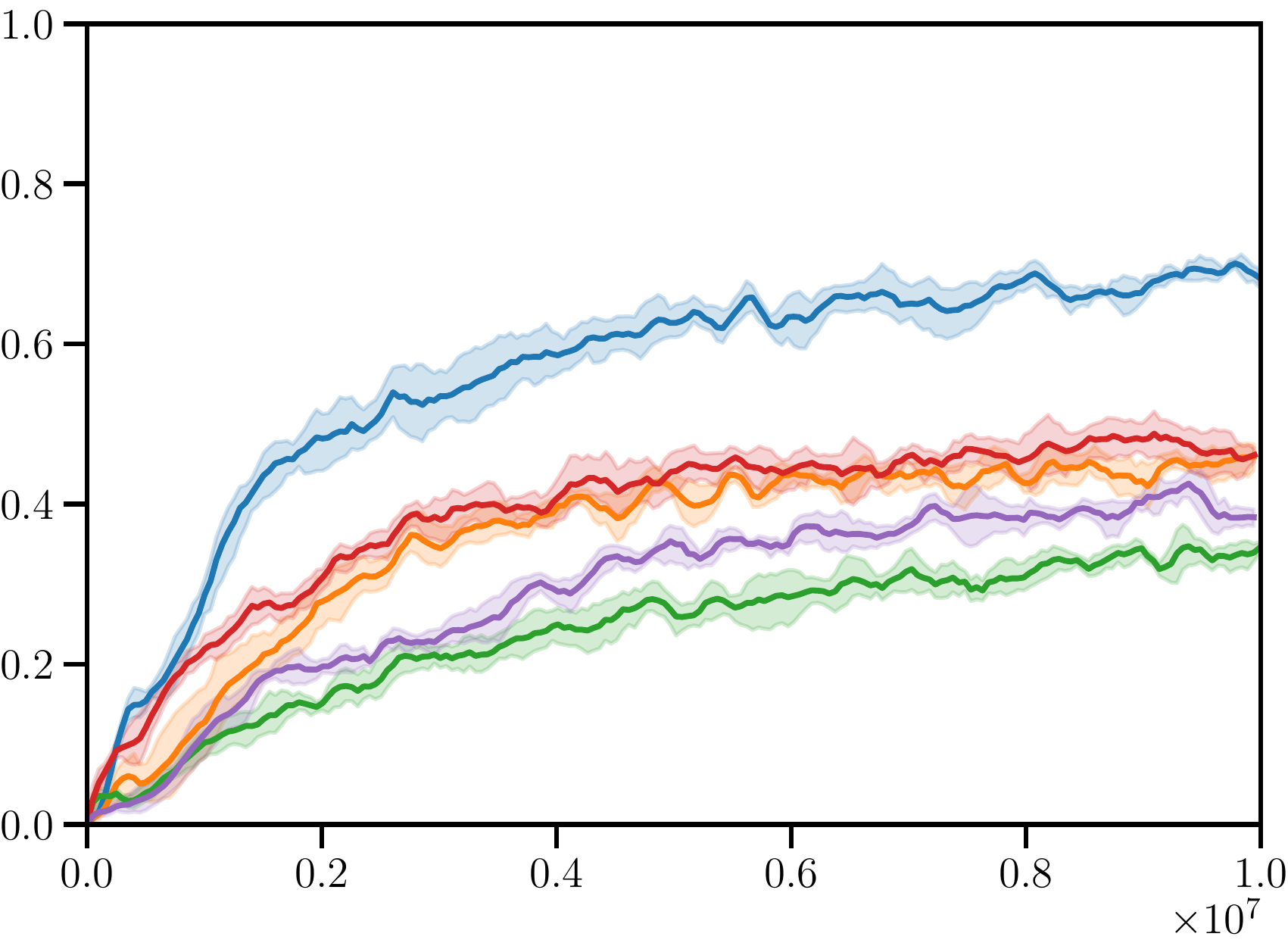}}
        \label{fSC_3-8MMM_win_rate}
    \end{subfigure}

    \begin{subfigure}[t]{0.333\textwidth}
        \centerline{\includegraphics[width=\linewidth]{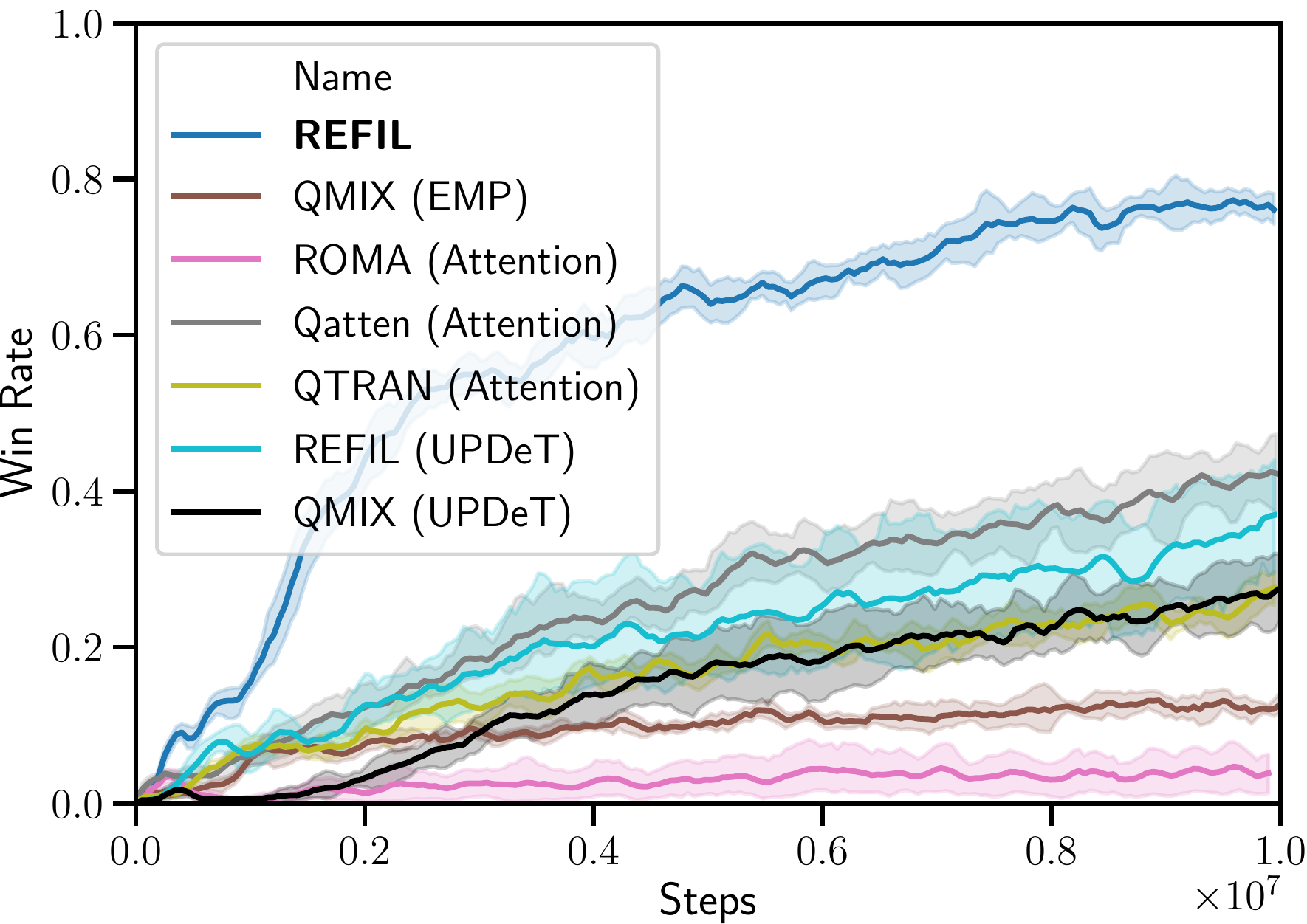}}
        \caption{3-8sz}
        \label{fSC_3-8sz_baselines_win_rate}
    \end{subfigure}
    \begin{subfigure}[t]{0.32\textwidth}
        \centerline{\includegraphics[width=\linewidth]{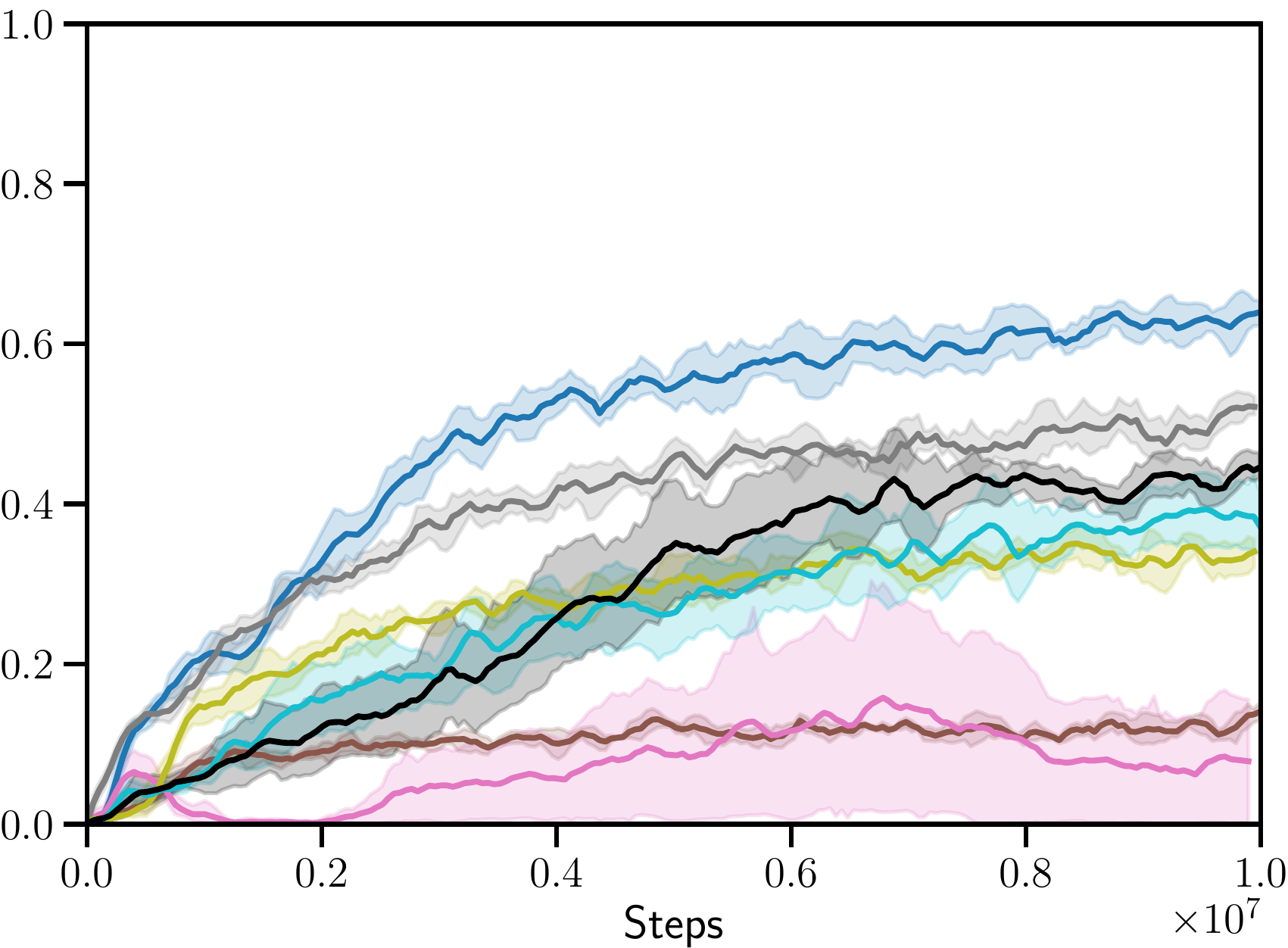}}
        \caption{3-8csz}
        \label{fSC_3-8csz_baselines_win_rate}
    \end{subfigure}
    \begin{subfigure}[t]{0.32\textwidth}
        \centerline{\includegraphics[width=\linewidth]{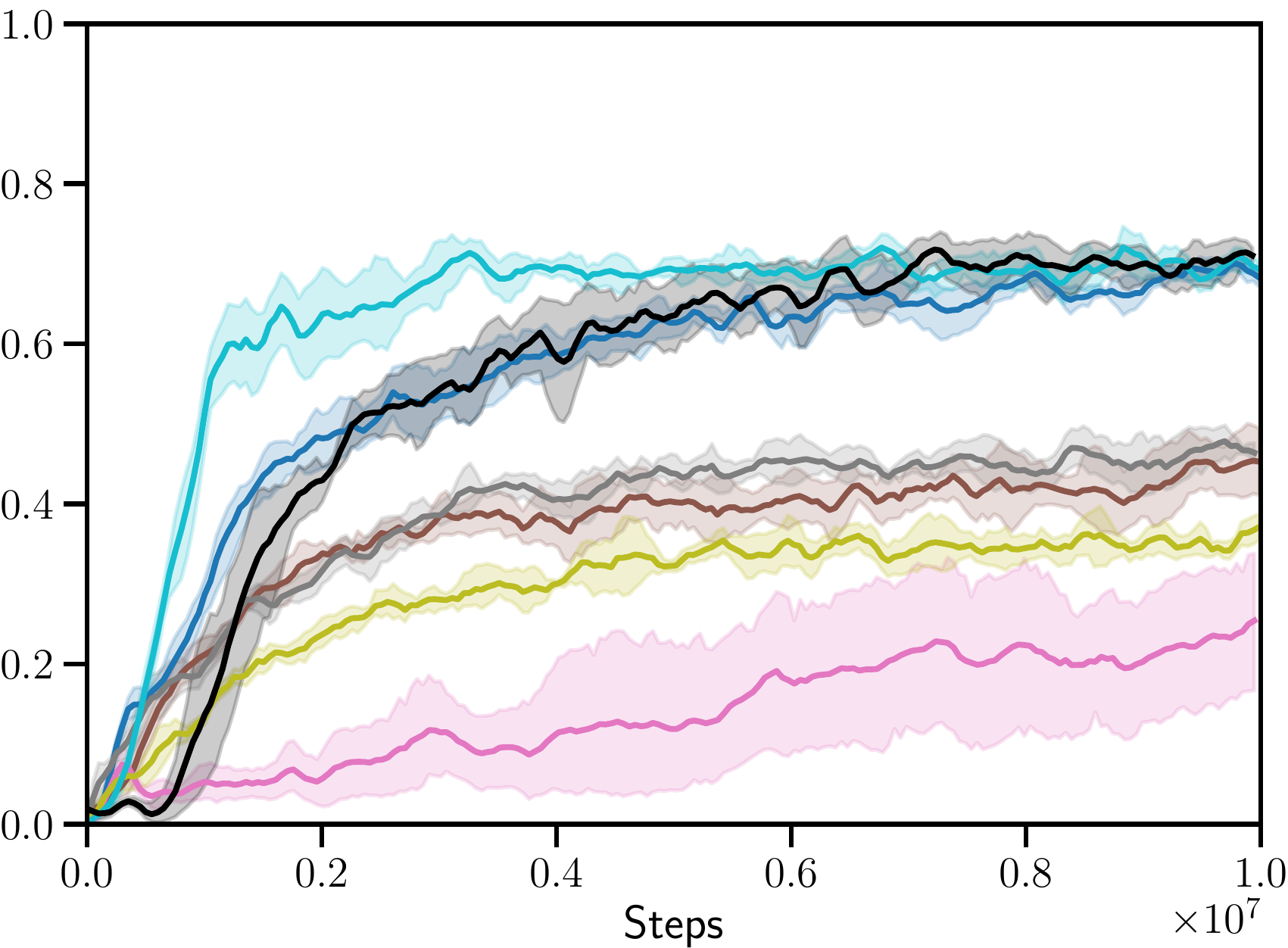}}
        \caption{3-8MMM}
        \label{fSC_3-8MMM_baselines_win_rate}
    \end{subfigure}
    \vspace{-0.1 in}
    \caption{\footnotesize
             Test win rate over time on multi-task \starcraft~environments.
             Tasks are sampled uniformly at each episode.
             Shaded region is a 95\% confidence interval across 5 runs.
             (top row) Ablations of our method.
             (bottom row) Baseline methods.
        }
    \label{fSC_results}
    \vspace{-0.2 in}
\end{figure*}

In our tests we evaluate on three settings we call 3-8sz, 3-8csz, and 3-8MMM.
3-8sz pits symmetrical teams of between 3 and 8 agents against each other where the agents are a combination of Zealots and Stalkers (inspired by the 2s3z and 3s5z tasks in the original SMAC), resulting in 39 unique tasks.
3-8csz pits symmetrical teams of between 0 and 2 Colossi and 3 to 6 Stalkers/Zealots against each other (inspired by 1c3s5z), resulting in 66 tasks.
3-8MMM pits symmetrical teams of between 0 and 2 Medics and 3 to 6 Marines/Marauders against each other (inspired by MMM and MMM2, again resulting in 66 tasks).

\begin{table}[t] 
    \vspace{-0.1 in}
    \scriptsize
    \captionsetup{width=\textwidth,justification=centering}
    \caption{Comparison of tested methods.}
    \vspace{-0.1 in}
    \centering
    \begin{tabular}{@{}llll@{}}
        \toprule
        \multirow{2}{*}{Name} & Imagined  & \multirow{2}{*}{Model} & Base  \\
                              & Learning  &                        & Algorithm \\
        \midrule
        \ourapproach       & \checkmark & MHA$^1$  & QMIX$^2$ \\
        QMIX (Attention)   &            & MHA      & QMIX     \\
        \ourapproach (VDN) & \checkmark & MHA      & VDN$^3$  \\
        VDN (Attention)    &            & MHA      & VDN      \\
        QMIX (Max Pooling) &            & Max-Pool & QMIX     \\
        \midrule
        QMIX (EMP)         &            & EMP$^4$  & QMIX     \\
        ROMA (Attention)   &            & MHA      & ROMA$^5$ \\
        Qatten (Attention) &            & MHA      & Qatten$^6$ \\
        QTRAN (Attention)  &            & MHA      & QTRAN$^7$ \\
        \ourapproach (UPDeT)& \checkmark& UPDeT$^8$& QMIX     \\
        QMIX (UPDeT)       &            & UPDeT    & QMIX     \\
        \bottomrule
    \end{tabular}
    \label{tab:baselines}
    \\
    $^1$: \citet{vaswani2017attention}
    $^2$: \citet{rashid2018qmix}
    $^3$: \citet{sunehag2017value} \\
    $^4$: \citet{agarwal2019learning}
    $^5$: \citet{wang2020roma}
    $^6$: \citet{yang2020qatten} \\
    $^7$: \citet{son2019qtran}
    $^8$: \citet{hu2021updet}
    \vspace{-0.1 in}
\end{table}

\textbf{Ablations and Baselines}\,
We introduce several ablations of our method, as well as adaptations of existing methods to handle variable sized inputs.
These comparisons are summarized in Table~\ref{tab:baselines}.
\textit{QMIX (Attention)} is our method without the auxiliary loss.
\ourapproach (VDN) is our approach using summation to combine all factors as in VDN rather than a non-linear monotonic mixing network.
\textit{VDN (Attention)} does not include the auxiliary loss and uses summation for factor mixing.
\textit{QMIX (Mean Pooling)} is \textit{QMIX (Attention)} with attention layers replaced by mean pooling.
We also test max pooling but find the performance to be marginally worse than mean pooling.
Importantly, for pooling layers we add entity-wise linear transformations prior to the pooling operations such that the total number of parameters is comparable to attention layers.

For baselines we consider some follow-up works to QMIX that improve the mixing network's expressivity: QTRAN~\citep{son2019qtran} and Qatten~\citep{yang2020qatten}.
We also compare to a method that builds on QMIX by attempting to learn dynamic roles that depend on the context each agent observes: ROMA~\citep{wang2020roma}.
We additionally consider an alternative mechanism for aggregating information across variable sets of entities, known as Entity Message Passing (EMP)~\citep{agarwal2019learning}.
We specifically use the restricted communication setting where agents can only communicate with agents they observe, and we set the number of message passing steps to three.
Finally, we consider the UPDeT architecture~\citep{hu2021updet}, a recent work that also targets the multi-task MARL setting.
UPDeT utilizes domain knowledge of the environment to map entities to the specific actions that they correspond to.
We train UPDeT with QMIX as well as \ourapproach.
For all approaches designed for the standard single-task SMAC setting, we extend them with the same multi-head attention architecture that our approach uses.

\textbf{Ablation Results}\,
Our results on challenges in multi-task \starcraft~ settings can be found in Figure~\ref{fSC_results}.
Tasks are sampled uniformly at each episode, so the curves represent average win rate across all tasks.
We find that \ourapproach outperforms all ablations consistently in these settings.
\textit{\ourapproach (VDN)} performs much worse than our approach and \textit{VDN (Attention)}, highlighting the importance of the mixing network handling contextual dependencies between entity partitions.
Since the trajectory of a subset of entities can play out differently based on the surrounding context, it is important for our factorization approach to recognize and adjust for these situations.
The use of mean-pooling in place of attention also performs poorly, indicating that attention is valuable for aggregating information from variable length sets of entities.

\begin{figure*}[t]
    \centering
    \sbox{\measurebox}{
        \begin{subfigure}[b]{0.78\textwidth}
            \includegraphics[width=\linewidth]{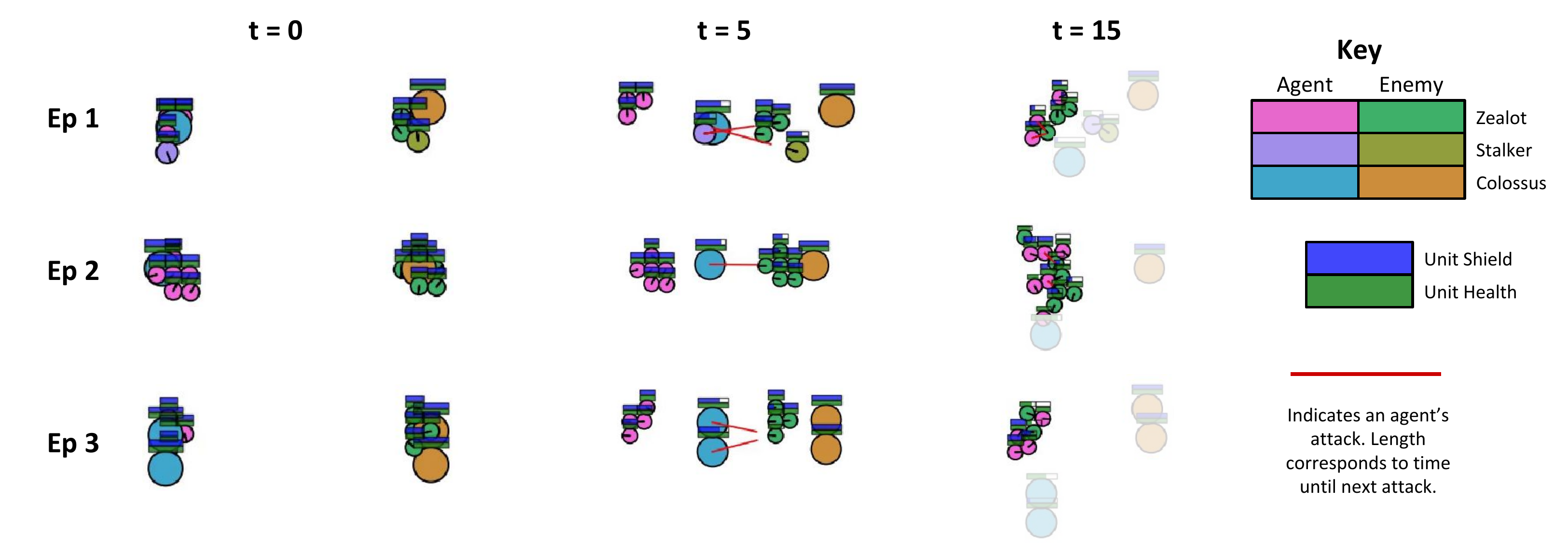}
        \end{subfigure}
    }
    \usebox{\measurebox}
    \begin{subfigure}[b][\ht\measurebox][s]{0.2\textwidth}
        \begin{subfigure}[b]{\linewidth}
            \includegraphics[width=\linewidth]{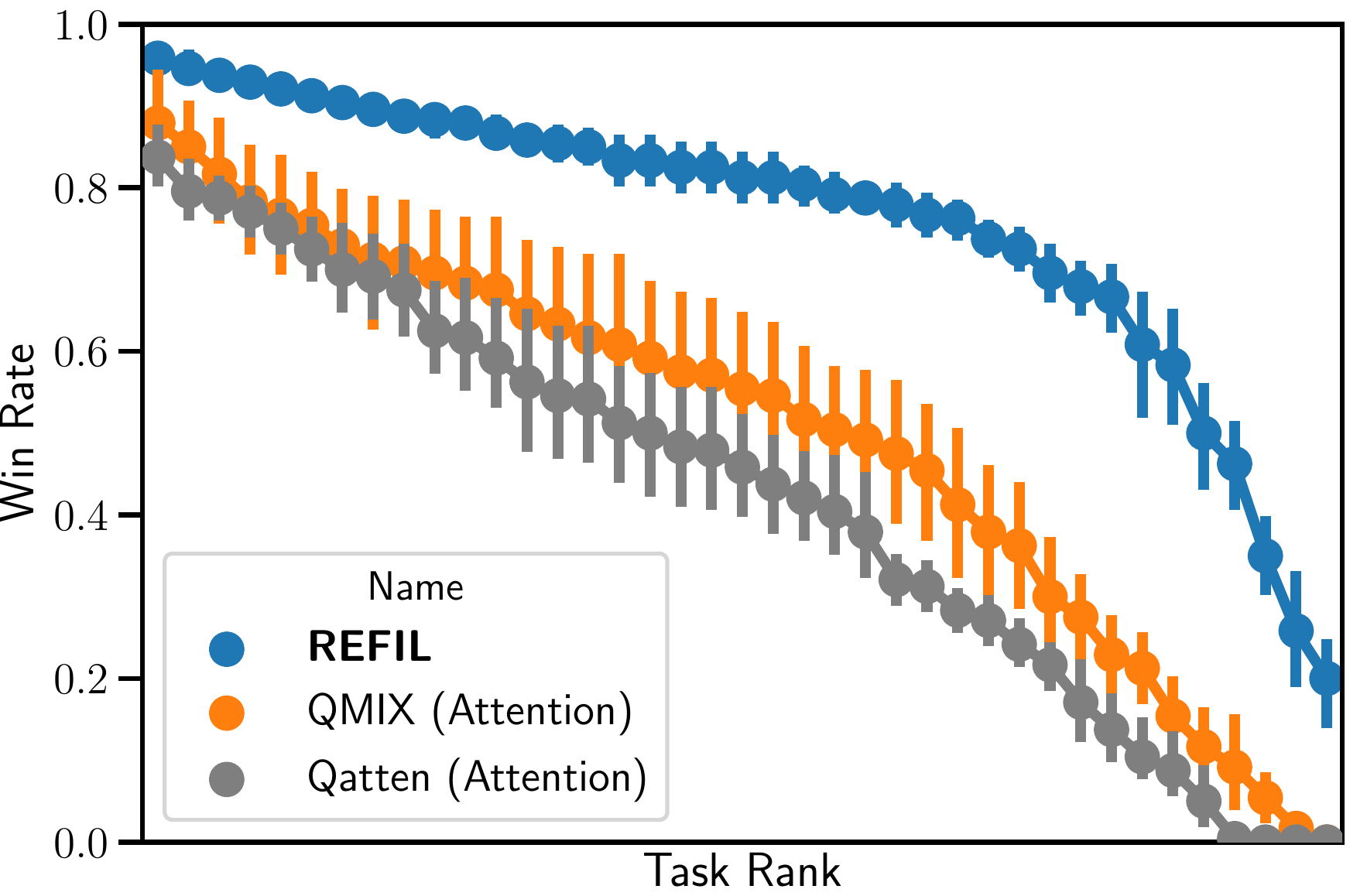}
        \end{subfigure}
        \begin{subfigure}[b]{\linewidth}
            \includegraphics[width=\linewidth]{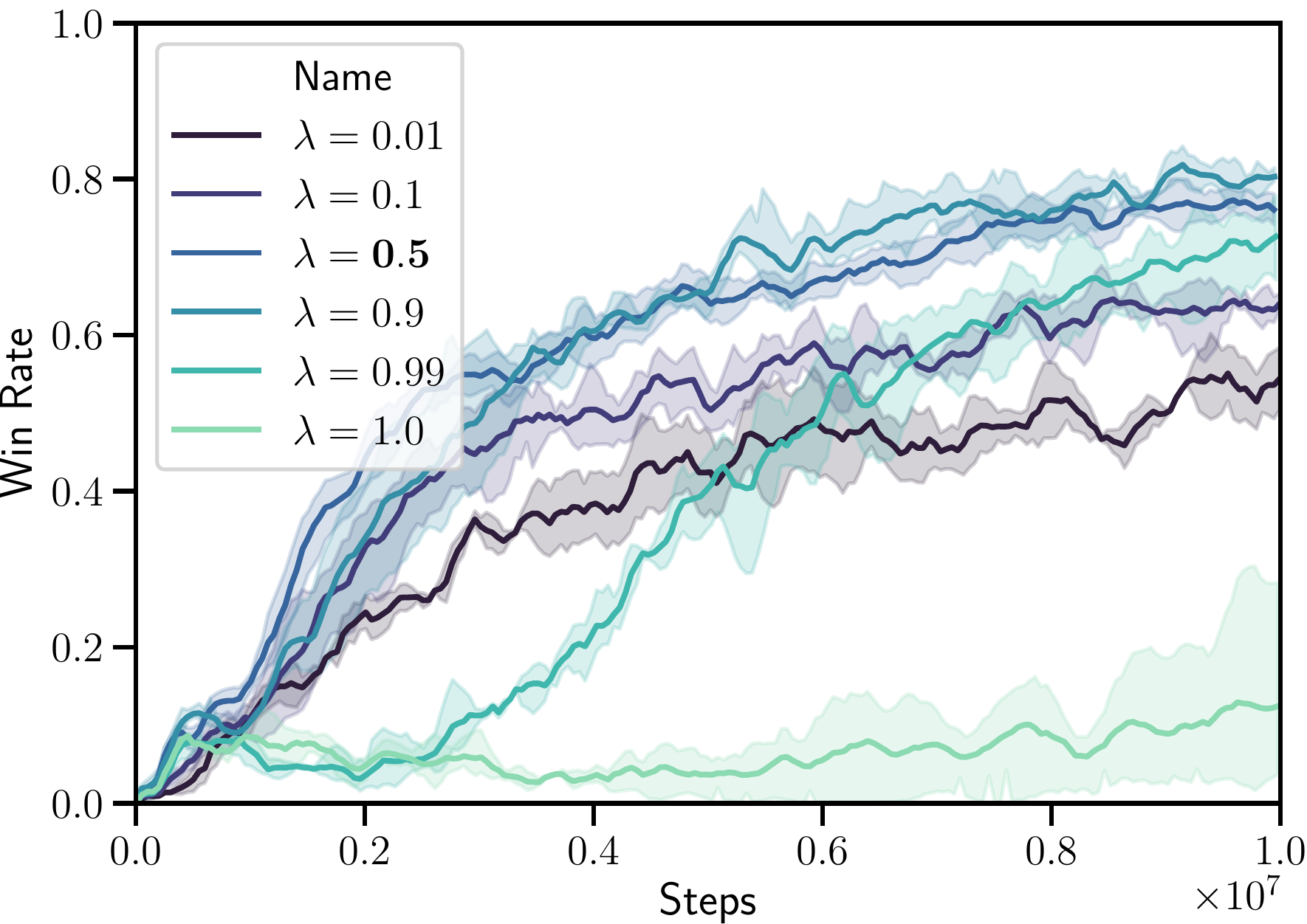}
        \end{subfigure}
    \end{subfigure}
    \vspace{-0.1 in}
    \caption{\footnotesize (left) Simplified rendering of a common pattern that emerges across tasks in the 3-8csz SMAC setting, highlighted at $t=15$.
    \ourapproach enables learning from each task to inform behavior in the others.
    (top right) Task-by-task performance on 3-8sz. \ourapproach generalizes better across a wider range of tasks.
    (bottom right) Varying $\lambda$ for \ourapproach in 3-8sz.
    }
    \label{fAdditionalResults}
    \vspace{-0.2 in}
\end{figure*}

\textbf{Baseline Results}\,
We find that algorithms designed to improve on QMIX for the single task MARL setting (ROMA, Qatten, QTRAN), naively applied to the multi-task setting, do not see the same improvements. \ourapproach, on the other hand, consistently outperforms other methods, highlighting the unique challenge of learning in multi-task settings.
In Fig.~\ref{fAdditionalResults} (top right) we investigate the performance of \ourapproach compared to the two next best methods in the 3-8sz setting on a task-by-task basis.
We evaluate each method on each task individually and rank the tasks by performance, plotting from left to right.
We find that the performance gain of \ourapproach comes from generalizing performance across a wider range of tasks, hence the reduced rate of decay in task performance from best to worst.

The entity aggregation method of EMP underperforms relative to the MHA module that we use.
UPDeT is a related work that focuses on designing an architecture compatible with multiple tasks and variable entities and action spaces by utilizing domain knowledge to map entities to their corresponding actions.
Despite adding this domain knowledge, QMIX (UPDeT) surprisingly underperforms in 2 of 3 settings, while performing similarly to \ourapproach on 3-8MMM; however, since UPDeT is an attention-based architecture, it is amenable to our proposed auxiliary training scheme.
We find applying random factorization to QMIX (UPDeT) improves its performance further in 3-8MMM as well as in 3-8sz.
In the case of 3-8MMM, where the asmyptotic win rate of \ourapproach (UPDeT) and QMIX (UPDeT) are similar, we find that \ourapproach (UPDeT) wins on average in 22\% fewer time steps by targeting enemy Medivacs, a unit capable of healing its teammates.
Targeting of Medivacs is an example of a common pattern that emerges across tasks which \ourapproach is able to leverage.

\textbf{Role of Auxiliary Objective}\,
In order to understand the role of training as an auxiliary objective (rather than entirely replacing the objective) we vary the value of $\lambda$ to interpolate between two modes: $\lambda = 0$ is simply \textit{QMIX (Attention)}, while $\lambda = 1$ trains exclusively with random factorization.
Our results on 3-8sz (Figure~\ref{fAdditionalResults} (bottom right)) show that, similar to regularization methods such as Dropout~\citep{srivastava2014dropout}, there is a sweet spot where performance is maximized before collapsing catastrophically.
Training exclusively with random factorization does not learn anything significant.
This failure is likely due to the fact that we use the full context in our targets for learning with imagined scenarios as well as when executing our policies, so we still need to learn with it in training.

\textbf{Qualitative Example of Common Pattern}\,
Finally, we visualize an example of the sort of common patterns that \ourapproach is able to leverage (Fig.~\ref{fAdditionalResults} (left)).
Zealots (the only melee unit present) are weak to Colossi, so they learn to hang back and let other units engage first.
Then, they jump in and intercept the enemy Zealots while all other enemy units are preoccupied, leading to a common pattern of a Zealot vs. Zealot skirmish (highlighted at t=15). 
\ourapproach enables behaviors learned in these types of sub-groups to be applied more effectively across all tasks.
By sampling groups from all entities randomly, we will occasionally end up with sub-groups that include only Zealots, and the value function predictions learned in these sub-groups can be applied not only to the task at hand, but to any task where a similar pattern emerges.

\vspace{-0.15 in}
\section{Related Work}
\label{sRelated}
\vspace{-0.1 in}
Multi-agent reinforcement learning (MARL) is a broad field encompassing cooperative~\citep{Foerster2017-do,rashid2018qmix,sunehag2017value}, competitive~\citep{Bansal2017-qm,Lanctot2017-gr}, and mixed~\citep{lowe2017multi,pmlr-v97-iqbal19a} settings.
This paper focuses on cooperative MARL with centralized training and decentralized execution \citep[CTDE]{oliehoek2016concise}.
Our approach utilizes value function factorization, an approach aiming to simultaneously overcome limitations of both joint \cite{hausknecht_cooperation_2016} and independent learning \cite{claus1998dynamics} paradigms.
Early attempts at value function factorisation require apriori knowledge of suitable per-agent team reward decompositions or interaction dependencies.
These include optimising over local compositions of individual $Q$-value functions learnt from individual reward functions  \citep{schneider_distributed_1999}, as well as summing individual $Q$-functions with individual rewards before greedy joint action selection \citep{russell_q-decomposition_2003}.
Recent approaches from cooperative deep multi-agent RL learn value factorisations from a single team reward function by treating all agents as independent factors, requiring no domain knowledge and enabling decentralized execution.
Value-Decomposition Networks (VDN) \citep{sunehag2017value} decompose the joint $Q$-value function into a sum of local utility functions used for greedy action selection. QMIX \cite{rashid2018qmix,JMLR:v21:20-081} extends such additive decompositions to general monotonic functions.
Some works extend QMIX to improve the expressivity of mixing functions~\citep{son2019qtran,yang2020qatten}, learn latent embeddings to help exploration~\citep{mahajan2019maven} or learn dynamic roles~\citep{wang2020roma}, and encode knowledge of action semantics into network architectures~\citep{wang2020action}.

Several recent works have addressed the topic of generalization and transfer across related tasks with varying agent quantities, though the learning paradigms considered and assumptions made differ from our approach.
\citet{carion2019structured} devise an approach for assigning agents to tasks, assuming the existence of low-level controllers to carry out the tasks, and show that it can scale to much larger tasks than those seen in training.
\citet{burden_deep_2020} propose a transfer learning approach using convolutional neural networks and grid-based state representations to scale to tasks of arbitrary size.
\citet{wang2021rode} introduce a method to decompose action spaces into roles, which they show can transfer to tasks with larger numbers of agents by grouping new actions into existing clusters.
They do not propose a model to handle the larger observation sizes, instead using a euclidean distance heuristic to observe a fixed number of agents.
Several approaches devise attention or graph-neural-network based models for handling variable sized inputs and focus on learning curricula to progress on increasingly large/challenging settings~\citep{Long2020Evolutionary,baker2019emergent,wang2020few,agarwal2019learning}.
Most recently, ~\citet{hu2021updet} introduce a method for handling variable-size inputs \emph{and} action spaces and evaluate their model on single-task to single-task transfer.
In contrast to these curriculum and transfer learning approaches, we focus on training simultaneously on multiple tasks and specifically develop a training paradigm for improving knowledge sharing across tasks.

\vspace{-0.15 in}
\section{Conclusion}
\label{sConclusion}
\vspace{-0.1 in}
In this paper we considered a multi-task MARL setting where we aim to learn control policies for variable-sized teams of agents.
We proposed \ourapproach, an approach that regularizes value functions to share factors comprised of sub-groups of entities, in turn promoting generalization and knowledge transfer within and across complex cooperative multi-agent tasks.
Our results showed that our contributions yield significant average performance improvements across these tasks when training on them concurrently, specifically through improving generalization across a wider variety of tasks.

\section*{Acknowledgements}
This work is partially supported by NSF Awards IIS-1513966/ 1632803/1833137, CCF-1139148, DARPA Award\#: FA8750-18-2-0117, FA8750-19-1-0504, DARPA-D3M - Award UCB-00009528, Google Research Awards, gifts from Facebook and Netflix, and ARO\# W911NF-12-1-0241 and W911NF-15-1-0484.
This project has also received funding from the European Research Council under the European Union's Horizon 2020 research and innovation programme (grant agreement number 637713).
The experiments were made possible by a generous equipment grant from NVIDIA.

We thank Greg Farquhar for developing the initial version of the StarCraft environment with dynamic team compositions.
We also thank Luisa Zintgraf, Natasha Jaques, and Robby Costales for their helpful feedback on the draft.

\bibliography{citations}
\bibliographystyle{icml2021}

\appendix
\onecolumn
\section{Attention Layers and Models}
\label{sMultiHeadAttention}
Attention models have recently generated intense interest due to their ability to incorporate information across large contexts.
Importantly for our purposes, they are able to process variable sized sets of inputs.

We now formally define the building blocks of our attention models. Given the input $\mX$, a matrix where the rows correspond to entities, we define an entity-wise feedforward layer as a standard fully connected layer that operates independently and identically over entities:
\begin{equation}
    \text{eFF}(\mX; \mW, \vb) = \mX \mW + \vb^\top, \mX \in \mathbb{R}^{n^x \times d} , \mW \in \mathbb{R}^{d \times h}, \vb \in \mathbb{R}^{h}
\end{equation}
Now, we specify the operation that defines an attention head, given the additional inputs of $\mathcal{S} \subseteq \mathbb{Z}^{[1, n^x]}$, a set of indices that selects which rows of the input $\mX$ are used to compute queries such that $X_{\mathcal{S},*} \in \mathbb{R}^{\abs{\mathcal{S}} \times d}$, and $\mM$, a binary obserability mask specifying which entities each query entity can observe (i.e. $\mM_{i,j} = 1$ when $i \in \mathcal{S}$ can incorporate information from $j \in \mathbb{Z}^{[1, n^x]}$ into its local context):
\begin{align}
    \label{eAtten}
    \text{Atten}(\mathcal{S}, \mX, \mat{M}; \mW^Q, \mW^K, \mW^V) = \text{softmax}\left( \text{mask} \left( \frac{\mat{Q} \mat{K}^\top}{\sqrt{h}}, \mM \right) \right) \mat{V} \in \mathbb{R}^{\abs{\mathcal{S}} \times h} \\
    \mat{Q} = \mX_{\mathcal{S},*} \mW^Q, \mat{K} = \mX \mW^K, \mat{V} = \mX \mW^V \,,\quad
    \mM \in \{0,1\}^{ \abs{\mathcal{S}} \times n^x}, \mW^Q, \mW^K, \mW^V \in \mathbb{R}^{d \times h}
\end{align}
The $\text{mask}(\mY, \mM)$ operation takes two equal sized matrices and fills the entries of $\mY$ with $-\infty$ in the indices where $\mM$ is equal to 0.
After the softmax, these entries become zero, thus preventing the attention mechanism from attending to specific entities.
This masking procedure is used in our case to uphold partial observability, as well as to enable ``imagining'' the utility of actions within sub-groups of entities.
Only one attention layer is permitted in the decentralized execution setting; otherwise information from unseen agents can be propagated through agents that are seen.
$\mW^Q$, $\mW^K$, and $\mW^V$ are all learnable parameters of this layer.
Queries, $\mat{Q}$, can be thought of as vectors specifying the type of information that an entity would like to select from others, while keys, $\mat{K}$, can be thought of as specifying the type of information that an entity possesses, and finally, values, $\mat{V}$, hold the information that is actually shared with other entities.

We define multi-head-attention as the parallel computation of attention heads as such:
\begin{equation}
    \text{MHA}\left( \mathcal{S}, \mX, \mat{M} \right) = \text{concat}\left( \text{Atten}\left( \mathcal{S}, \mX, \mat{M}; \mW_j^Q, \mW_j^K, \mW_j^V \right), j \in \left( 1 \dots n^h \right) \right)
\end{equation}
The size of the parameters of an attention layer does not depend on the number of input entities.
Furthermore, we receive an output vector for each query vector.

\section{Augmenting QMIX with Attention}
\label{sAttenQMIX}
The standard QMIX algorithm relies on a fixed number of entities in three places: inputs of the agent-specific utility functions $Q_a$, inputs of the hypernetwork, and the number of utilities entering the mixing network, which must correspond the output of the hypernetwork since it generates the parameters of the mixing network.
QMIX uses multi-layer perceptrons for which all these quantities have to be of fixed size.
In order to adapt QMIX to the variable agent quantity setting, such that we can apply a single model across all tasks, we require components that accept variable sized sets of entities as inputs.
By utilizing attention mechanisms, we can design components that are no longer dependent on a fixed number of entities taken as input.
We define the following inputs: $\bm{X}_{ei}^{\mathcal E} := s^e_i, 1 \leq i \leq d, e \in \mathcal E; \bm{M}_{ae}^\mu := \mu(s^a, s^e), a \in \mathcal A, e \in \mathcal E$.
The matrix $\mX^\mathcal{E}$ is the global state $\mathbf{s}$ reshaped into a matrix with a row for each entity, and $\mM^{\mu}$ is a binary observability matrix which enables decentralized execution, determining which entities are visible to each agent.

\subsection{Utility Networks}
While the standard agent utility functions map a flat observation, whose size depends on the number of entities in the environment, to a utility for each action, our attention-utility functions can take in a variable sized set of entities and return a utility for each action.
The attention layer output for agent $a$ is computed as $\text{MHA}\left( \{ a \}, \mX, \mat{M}^{\mu} \right)$, where $\mX$ is an row-wise transformation of $\mX^\mathcal{E}$ (e.g., an entity-wise feedforward layer).
If agents share parameters, the layer can be computed in parallel for all agents by providing $\mathcal{A}$ instead of $\{ a \}$, which we do in practice.

\subsection{Generating Dynamic Sized Mixing Networks}
Another challenge in devising a QMIX algorithm for variable agent quantities is to adapt the hypernetworks that generate weights for the mixing network.
Since the mixing network takes in utilities from each agent, we must generate feedforward mixing network parameters that change in size depending on the number of agents present, while incorporating global state information.
Conveniently, the number of output vectors of a MHA layer depends on the cardinality of input set $\mathcal S$ and we can therefore generate mixing parameters of the correct size by using $\mathcal S = \mathcal A$ and concatenating the vectors to form a matrix with one dimension size depending on the number of agents and the other depending on the number of hidden dimensions.
Attention-based QMIX (QMIX (Attention)) trains these models using the standard 
DQN loss in Equation 2 of the main text.

Our two layer mixing network requires the following parameters to be generated: $\mW_1 \in \mathbb{R}^{+(\abs{\mathcal{A}} \times h^m)}$, $\vct{b}_1 \in \mathbb{R}^{h^m}$, $\vct{w}_2 \in \mathbb{R}^{+(h^m)}$, $b_2 \in \mathbb{R}$, where $h^m$ is the hidden dimension of the mixing network and $\abs{\mathcal{A}}$ is the set of agents.

Note from Eq.~\eqref{eAtten} that the output size of the layer is dependent on the size of the query set.
As such, using attention layers, we can generate a matrix of size $\abs{\mathcal{A}} \times h^m$, by specifying the set of agents, $\mathcal{A}$, as the set of queries $\mathcal{S}$ from Eq.~\eqref{eAtten}.
We do not need observability masking since hypernetworks are only used during training and can be fully centralized.
For each of the four components of the mixing network ($\mW_1, \vct{b}_1, \vct{w}_2, b_2$), we introduce a hypernetwork that generates parameters of the correct size. 
Thus, for the parameters that are vectors ($\vct{b}_1$ and $\vct{w}_2$), we average the matrix generated by the attention layer across the $\abs{\mathcal{A}}$ sized dimension, and for $b_2$, we average all elements.
This procedure enables the dynamic generation of mixing networks whose input size varies with the number of agents.
Assuming $\vq = [Q^1(\tau^1, u^1), \dots, Q^n(\tau^n, u^n)]$, then $\Qtot$ is computed as:
\begin{equation}
    \label{eMixingNet}
    \Qtot(\mathbf{s}, \mathbf{\tau}, \mathbf{u}) = \sigma((\vq^\top \mW_1) + \vct{b}_1^\top) \vct{w}_2 + b_2
\end{equation}
where $\sigma$ is an ELU nonlinearity~\citep{clevert2015fast}.

\section{Environment Details}
\subsection{\starcraft with Variable Agents and Enemies}
The standard version of SMAC loads map files with pre-defined and fixed unit types, where the global state and observations are flat vectors with segments corresponding to each agent and enemy.
Partial observability is implemented by zeroing out segments of the observations corresponding to unobserved agents.
The size of these vectors changes depending on the number of agents placed in the map file.
Furthermore, the action space consists of movement actions as well as separate actions to attack each enemy unit.
As such the action space also changes as the number of agents changes.

Our version loads empty map files and programmatically generates agents, allowing greater flexibility in terms of the units present to begin each episode.
The global state is split into a list of equal-sized entity descriptor vectors (for both agents and enemies), and partial observability is handled by generating a matrix that shows what entities are visible to each agent.
The variable-sized action space is handled by randomly assigning each enemy a tag at the beginning of each episode and designating an action to attack each possible tag, of which there are a maximum number (i.e. the maximum possible number of enemies across all tasks).
Agents are able to see the tag of the enemies they observe and can select the appropriate action that matches this tag in order to attack a specific enemy.
Since UPDeT~\citep{hu2021updet} naturally handles the variable-sized action space and the mapping of entities to actions, we ensure that the output action utilities for UPDeT are shuffled based on the tags to maintain the entity-to-action mapping.

\section{Experimental Details}
Our experiments were performed on a desktop machine with a 6-core Intel Core i7-6800K CPU and 3 NVIDIA Titan Xp GPUs, and a server with 2 16-core Intel Xeon Gold 6154 CPUs and 10 NVIDIA Titan Xp GPUs.
Each experiment is run with 8 parallel environments for data collection and a single GPU.
\ourapproach takes about 24 hours to run for 10M steps on \starcraft.
QMIX (Attention) takes about 16 hours for the same number of steps on \starcraft.
Reported times are on the desktop machine and the server runs approximately 15\% faster due to more cores being available for running the environments in parallel.

\section{Hyperparameters}
Hyperparameters were based on the PyMARL~\citep{samvelyan2019starcraft} implementation of QMIX and are listed in Table~\ref{tab:hyperparams}.
All hyperparameters are the same in all \starcraft settings.
Since we train for 10 million timesteps (as opposed to the typical 2 million in standard SMAC), we extend the epsilon annealing period (for epsilon-greedy exploration) from 50,000 steps to 500,000 steps.
For hyperparameters new to our approach (hidden dimensions of attention layers, number of attention heads, $\lambda$ weighting of imagined loss), the specified values in Table~\ref{tab:hyperparams} were the first values tried, and we found them to work well.
The robustness of our approach to hyperparameter settings, as well as the fact that we do not tune hyperparameters per environment, is a strong indicator of the general applicability of our method.

\begin{table*}[ht]
\captionsetup{width=0.8\textwidth,justification=centering}
\begin{center}
    \caption{Hyperparameter settings across all runs and algorithms/baselines.}
    \begin{tabular}{@{}lll@{}}
        \toprule
        Name & Description & Value \\
        \midrule
        lr & learning rate & $0.0005$ \\
        optimizer & type of optimizer & RMSProp$^1$ \\
        optim $\alpha$ & RMSProp param & $0.99$ \\
        optim $\epsilon$ & RMSProp param & $1e-5$ \\
        target update interval & copy live params to target params every \textunderscore~ episodes & $200$ \\
        bs & batch size (\# of episodes per batch) & $32$ \\
        grad clip & reduce global norm of gradients beyond this value & $10$ \\
        \midrule
        $|D|$ & maximum size of replay buffer (in episodes) & $5000$ \\
        $\gamma$ & discount factor & $0.99$ \\
        starting $\epsilon$ & starting value for exploraton rate annealing & $1.0$ \\
        ending $\epsilon$ & ending value for exploraton rate annealing & $0.05$ \\
        anneal time & number of steps to anneal exploration rate over & $500000$ \\
        \midrule
        $h^a$ & hidden dimensions for attention layers & 128 \\
        $h^r$ & hidden dimensions for RNN layers & 64 \\
        $h^m$ & hidden dimensions for mixing network & 32 \\
        \# attention heads & Number of attention heads & 4 \\
        nonlinearity & type of nonlinearity (outside of mixing net) & ReLU \\
        \midrule
        $\lambda$ & Weighting between standard QMIX loss and imagined loss & $0.5$ \\
        \bottomrule
    \end{tabular}
    \label{tab:hyperparams} \\
    $^1$: \citet{tieleman2012lecture}
\end{center}
\end{table*}

\end{document}